\documentclass[10pt,journal,compsoc]{IEEEtran}
\pdfoutput=1
\usepackage{siunitx}
\ifCLASSOPTIONcompsoc
  \usepackage[nocompress]{cite}
\else
  \usepackage{cite}
\fi
\usepackage{verbatim}

\ifCLASSINFOpdf
\usepackage{graphicx}
\usepackage[FIGTOPCAP]{subfigure}
\usepackage{hyperref}
\usepackage[export]{adjustbox}
\usepackage{blindtext}

\usepackage{amsmath}
\DeclareMathOperator*{\argmin}{argmin}
\usepackage{mathtools}  
\usepackage{array}
\usepackage{fixltx2e}
\usepackage{stfloats}
\usepackage{url}
\usepackage{footnote}

\usepackage{enumitem}
\usepackage{booktabs} % For formal tables
\usepackage{placeins}
\usepackage{multirow}
\usepackage{array}
\newcolumntype{L}[1]{>{\raggedright\let\newline\\\arraybackslash\hspace{0pt}}m{#1}}

\begin{document}
\title{Adaptation Strategies for Automated Machine Learning on Evolving Data}

\author{Bilge Celik
        and
        Joaquin Vanschoren
        % <-this % stops a space
\IEEEcompsocitemizethanks{\IEEEcompsocthanksitem B. Celik and J. Vanschoren work at the Eindhoven University of Technology, Eindhoven, the Netherlands.\protect\\
E-mail: B.Celik.Aydin@tue.nl}% <-this % stops an unwanted space
\thanks{Manuscript received ...}}



\IEEEtitleabstractindextext{%
\begin{abstract}
Automated Machine Learning (AutoML) systems have been shown to efficiently build good models for new datasets. However, it is often not clear how well they can adapt when the data evolves over time. The main goal of this study is to understand the effect of concept drift on the performance of AutoML methods, and which adaptation strategies can be employed to make them more robust to changes in the underlying data. To that end, we propose 6 concept drift adaptation strategies and evaluate their effectiveness on a variety of AutoML approaches for building machine learning pipelines, including Bayesian optimization, genetic programming, and random search with automated stacking. These are evaluated empirically on real-world and synthetic data streams with different types of concept drift. Based on this analysis, we propose ways to develop more sophisticated and robust AutoML techniques.
\end{abstract}

\begin{IEEEkeywords}
AutoML, data streams, concept drift, adaptation strategies
\end{IEEEkeywords}}

\maketitle

\IEEEdisplaynontitleabstractindextext

% \fi
\IEEEpeerreviewmaketitle

\IEEEraisesectionheading{\section{Introduction}\label{sec:introduction}}
\IEEEPARstart{T}{he} field of automated machine learning (AutoML) aims to automatically design and build machine learning systems, replacing manual trial-and-error with systematic, data-driven decision making \cite{Yao2018}. This makes robust, state-of-the-art machine learning accessible to a much broader range of practitioners and domain scientists.

Although AutoML has been shown to be very effective and even rival human machine learning experts \cite{Thornton2013,Feurer2015,Olson2016,H2O2019}, they are usually only evaluated on static datasets \cite{Yao2018}. However, data is often not static, it evolves. As a result, models that have been found to work well initially may become sub-optimal as the data changes over time. Indeed, most AutoML techniques assume that earlier evaluations are forever representative of new data, and may therefore fail to adapt to changes in the underlying distribution of the data, also known as concept drift \cite{Gomes2017,Oza2001,Bifet2010}. 

There has been very limited work on automating the design of machine learning pipelines in settings with concept drift. That would require that the AutoML process, the automated search for optimal pipelines, is adapted so that it can cope with concept drift and adjust the pipelines over time, whenever the data changes so drastically that a pipeline redesign or re-tuning is warranted.
Simply re-running AutoML techniques from scratch on new data is often not feasible, since AutoML techniques are usually computationally expensive, and in most online learning settings good predictive performance must be achieved in a limited time window.

To truly understand the effects of concept drift and how to cope with it in AutoML systems, we systematically study how different AutoML techniques are affected by different types of concept drift. We discover that, while these AutoML methods often perform very similarly on static datasets \cite{Gijsbers2019b}, they each respond very differently to different types of concept drift. We propose six different adaptation strategies to cope with concept drift and implement these in open-source AutoML libraries. We find that these can indeed be effectively used, paving the way towards novel AutoML techniques that are robust against evolving data. 

The online learning setting under consideration is one where data can be buffered in \textit{batches} using a moving window. Hence, we assume limited memory and limited time for making new predictions, and evaluate the performance of different adaptation strategies paired with different AutoML methods accordingly. Online settings where only one new instance fits in memory and can only be viewed once (single pass) are beyond the scope of this paper.

%\textit{This paper focuses on adaptation of current AutoML methods to different concept drifts in evolving data and guide future research on design of an automated stream learning system. Therefore, the underlying constraints are somewhat different than classical online learning settings. A common approach of buffering data in batches is taken in order to give more time to re-optimize AutoML, as well as relaxation of the limited memory and single pass constraints.}
%NEW UP

The paper is organized as follows. First, we formalize the problem in Section \ref{sec:problem}. Section \ref{sec:rwork} covers related work as well as the most relevant AutoML systems and online learning techniques. Our adaptation strategies for AutoML methods are described in Section \ref{sec:adaptation}. Section \ref{sec:experiments} details our empirical setup to evaluate these strategies, and the results are analyzed in Section \ref{sec:results}. Section \ref{sec:conclusions} concludes.  

\section{Problem Definition}\label{sec:problem}
%In this section, we formalize AutoML and concept drift.

\subsection{AutoML}
AutoML methods aim to design an optimal series of operations that transform raw data into desired outputs, such as a machine learning pipeline or a neural architecture \cite{Vanschoren2018}. In this paper, we are concerned with optimizing machine learning pipelines, which can be formally defined as a \textit{combined algorithm selection and hyperparameter optimization (CASH)} problem \cite{Thornton2013}. 
Given a training set $ X_{tr} $ with targets $ y_{tr} $, and a set of algorithms $ A = \big\{A^1, ... ,A^p\big\}$, each with a space of hyperparameters $ \big\{\Lambda^1, ..., \Lambda^p\big\}$, the CASH problem is defined as finding the algorithm $A^*$ and hyperparameter settings $\lambda$ that minimize a loss function $ L $ (e.g. misclassification loss) applied to a model trained on $\big\{X_{tr}, y_{tr}\} $, and evaluated on a validation set $ \big\{X_{val}, y_{val}\} $, e.g. with \textit{k}-fold cross validation where $ \big\{X^i, y^i\} $ is a subset of the data set:
\begin{equation}
\begin{split}
A^*_{\lambda} \in
\argmin_{%
       \substack{%
         \forall A^* \in A \\
         \forall \lambda \in \Lambda
       }
     }
\frac{1}{k} 
\sum_{i=1}^{k} L(A^j_\lambda, \big\{X^i_{tr}, y^i_{tr}\big\}, \big\{X^i_{val}, y^i_{val}\big\})
\end{split}
\label{eq:1}
\end{equation}

Instead of a single learning algorithm, $A$ can be a full pipeline including data preprocessing, feature preprocessing, meta-learning and model post-processing, which typically creates a large search space of possible pipelines and configurations. This results in important trade-offs between computational efficiency (e.g. training and prediction time) and accuracy \cite{Yao2018}.

\subsection{Online learning and concept drift}
%definition, types, example, online drift adaptation, dynamic learning, existing most well known approaches
In online learning, models are trained and used for prediction without having all training data beforehand. The data stream should be considered to be infinitely long, has to be processed in order, and only a limited part of it can be in memory at any given time \cite{Gama2014}. Oftentimes, the data points are processed in small batches, also called \textit{windows}. The process that generates the data may change over time, leading to \textit{concept drift}, a usually unpredictable shift over time in the underlying distribution of the data.

Consider a data stream $\big\{ X, y\big\}$ generated from a joint probability density function $p(X, y)$, also called the \textit{concept} that we aim to learn. Concept drift can be described as:

\begin{equation}
\exists X: p_{t0} (X, y) \neq p_{t1} (X, y)
\end{equation}

where $p_{t0}(X,y)$ and $p_{t1}(X,y)$ represent joint probability functions at time $t_0$ and $t_1$, respectively \cite{Gama2014}. Webb et al. \cite{Webb2016} further categorize concept drift into 9 main classes based on several quantitative and qualitative characteristics, of which the \textit{duration} and \textit{magnitude} of the drift have the greatest impact on learner selection and adaptation. \textit{Drift duration} is the amount of time in which an initial concept ($a_0$, at time $t_0$) drifts to a resulting concept ($a_1$, at time $t_1$):

\begin{equation}
Duration(a_0, a_1) = t_1 - t_0
\end{equation}

\textit{Abrupt (sudden) drift} is a change in concept occurring within a small time window $\gamma$: 

\begin{equation}
Abrupt Drift \Rightarrow Duration(a_0, a_1) < \gamma
\end{equation}

\textit{Drift magnitude} is the distance between the initial and resulting concepts over the drift period $[t_0,t_1]$: 

\begin{equation}
Magnitude(t_0, t_1) = D(a_0, a_1)
\end{equation}

where $D$ is a distribution distance function that quantifies the difference between concepts at two points in time. Webb et al. \cite{Webb2016} argue that magnitude will have a great impact on the ability of a learner to adapt to the drift. A minor abrupt drift may require refining a model, whereas major abrupt drift may require abandoning the model completely.

In \textit{gradual drift}, the drift magnitude over a time period is smaller than a maximum difference $\mu$ between the concepts:

\begin{equation}
Gradual Drift \Rightarrow \forall _{t \in [t_0, t_1]} D(a_0, a_1) < \mu
\end{equation}

%Most current AutoML solutions are designed for offline learning and have to be adapted to be used in an online setting with concept drift, which poses significant new challenges. 
It is crucial to understand the dynamics of concept drift and its effect on the search strategy used by the AutoML technique in order to design a successful adaptation strategy. Drift detection algorithms (e.g. DDM \cite{Gama2004}) are a key part of these strategies. They determine the location of drifts to alarm the learner so that it can react in a timely manner.

\section{Related Work}\label{sec:rwork}
To the best of our knowledge, there has been very little work that aims to understand how AutoML techniques can be adapted to deal with concept drift, even though there is clear evidence that most current AutoML techniques are greatly affected by it. A recent AutoML challenge focused on concept drift and found that current AutoML approaches did not adapt to concept drift in limited time frames, and were outperformed by incremental learning techniques, especially gradient boosting, with clever preprocessing \cite{escalante2020automl}.

There exists interesting work on speeding up hyperparameter optimization by transfer learning from prior tasks \cite{Golovin2017}, or continual learning \cite{de2019continual} where the goal is to adapt (deep learning) models to new tasks without catastrophic forgetting. However, these do not consider concept drift in single tasks, and evolving data cannot always easily be split up into multiple tasks. In the online learning literature there exist many techniques for handling different types of concept drift \cite{Gama2014}, but these usually adapt a previously chosen model rather than redesigning entire machine learning pipelines, as is the goal here. Interesting early results in algorithm selection were obtained using meta-learning to recommend algorithms over different streams \cite{gama2011learning}. More recently, hyperparameter tuning techniques have been proposed which re-initiate hyperparameter tuning when drift is detected \cite{veloso2018self,carnein2019towards}. However, these are tied to specific optimization techniques for single algorithms, while we aim to generally adapt AutoML techniques to redesign or re-optimize entire pipelines. There also exist strategies to adapt models previously learned by online learning algorithms to new data, usually through ensembling \cite{Bakirov2018}, but these do not re-optimize the algorithms or pipelines themselves.

There is some interesting work in optimization that could potentially be leveraged. For instance, there exists work that enables Bayesian models to detect changepoints in the optimization process \cite{garnett2010learning, garnett2010sequential}, which could potentially be used to adapt the Bayesian Optimization techniques used in certain AutoML methods, but to the best of our knowledge this has not been previously explored. 

Most similar to our work is a recent study by Madrid et al. \cite{Madrid2019}, in which a specific AutoML method (autosklearn) is extended with concept drift detection and two model adaptation methods. In this paper, however, we extend not one, but a range of very different AutoML techniques. We also generate datasets with very different types of concept drift to be able to understand how these AutoML approaches are each affected in their own way, as well as how quickly they recover from concept drift, if at all. In addition, we propose and evaluate five different adaptation techniques that radically differ in their resource requirements and how they train candidate models. All of this provides clear empirical evidence and guidance on how to develop novel AutoML techniques more suited to evolving data.

In the remainder of this section, we will discuss the AutoML and online learning techniques used in this study.

\subsection{AutoML techniques and systems}
%give a summary of all automl systems, justify selection of 3, then detailed explanation of following in subsections.
\textit{Bayesian Optimization (BO)} is one of the most successfully used AutoML approaches in the literature \cite{Brochu2010}. To efficiently explore the large space of possible pipeline configurations, it trains a probabilistic \textit{surrogate model} on the previously evaluated configurations to predict which unseen configurations should be tried next. The trade-off in this process is between exploitation of currently promising configurations versus exploration of new regions. In \textit{Sequential Model-Based Optimization (SMBO)} configurations are evaluated one by one, each time updating the surrogate model and using the updated model to find new configurations. Popular choices for the surrogate model are \textit{Gaussian Processes}, shown to give better results on problems with fewer dimensions and numerical hyperparameters \cite{Snoek2012}, whereas Random Forest-based approaches are more successful in high-dimensional hyperparameter spaces of a discrete nature \cite{Feurer2015}. \textit{Tree-structured Parzen Estimators} (TPE) are more amenable to parallel evaluation of configurations \cite{Bergstra2011}.

\textit{Evolutionary computation} offers a very different approach. For instance, pipelines can be represented as trees and genetic programming can be used to cross-over and/or mutate the best pipelines to evolve them further, growing in complexity as needed \cite{Olson2016}. \textit{Random search} also remains a popular technique. While less sample-efficient, it can be easily parallellized and/or combined with other strategies. 

General strategies to improve AutoML include multi-fidelity optimization techniques, which first try many configurations on small samples of the data, only evaluating the best ones on more data. Ensembling techniques such as voting or stacking can combine many previously trained configurations, and correct for over- or underfitting. Finally, it is often beneficial to use \textit{meta-learning} to build on information gained from previous experiments, for instance to \textit{warm-start} the search with the most promising configurations, or to design a smaller search space. For a wide survey of AutoML techniques, see \cite{Vanschoren2018}.

In this paper, we will study the impact of concept drift on each of these approaches (BO, evolution, and random search). Since we need to adapt these approaches with novel adaptation strategies to handle concept drift, we select open-source, state-of-the-art AutoML systems for each of these.

\subsubsection{Autosklearn}
Autosklearn \cite{Feurer2015} is an AutoML system using Bayesian optimization (SMBO). It produces pipelines consisting of a wide range of classifiers and pre-processing techniques from scikit-learn \cite{scikit-learn2011}. It supports warm-starting by initiating the search with pipelines that worked well on similar prior data sets, as well as a greedy ensemble selection technique to build ensembles out of the different pipelines tried.

\subsubsection{H2O AutoML}
H2O AutoML performs random search in combination with automated stacked ensembles \cite{H2O2019}. The stacking technique can be Random Forests, Gradient Boosting Machines (GBM), Deep Neural Nets, or generalized linear models (GLM). 
%The default is GLM stacking over a fixed grid of configurations. 
The stacked ensembles can contain either the best of all base models, or the best model from each algorithm family.

\subsubsection{GAMA}
GAMA uses genetic programming to automatically generate machine learning pipelines \cite{Gijsbers2019}. It is similar to TPOT \cite{Olson2016}, but uses asynchronous rather than synchronous evolution, and includes automated ensembling as well.

\subsection{Learning on evolving data}
In order to evaluate AutoML techniques on evolving data, we build on best practices in online learning.

\subsubsection{Forgetting mechanisms}
In order to adapt better to the changing environment, irrelevant past data should be forgotten. A popular forgetting mechanisms is a sliding window where data is disregarded at a constant rate \cite{Gama2014}. Alternatively, data can be down-weighted with a dynamic rate when concept drift is detected, or removed based on the class distribution. The most appropriate forgetting mechanism depends on the environment and drift characteristics. In this work we will explore strategies using both sliding windows and concept drift detection.

%Other distinctive characteristics of online learning are the limitation of a single pass of the algorithm through data and the necessity of incremental learning where data should not be required as a whole at the beginning \cite{Gaber2007}. In this study, we modify AutoML methods to incorporate these properties and track relevant performance measures. 

\subsubsection{Online learning algorithms}
Although many stream learning algorithms use a single learner (e.g. Hoeffding Trees \cite{Domingos2000}), ensemble learners (e.g. SEA \cite{Street2001}, Blast \cite{vanRijn2016}) are commonly used because of their higher accuracy and faster adaptation to concept drift.
%Online ensembles are dynamic learners where classifiers are added incrementally with a single sample at a time. 
Oza's online version of bagging \cite{Oza2001} performs a majority vote over the base models. Leveraging Bagging \cite{Bifet2010} adds more input and output randomization to the base models to increase ensemble diversity at the cost of computational efficiency. Blast \cite{vanRijn2016} builds a heterogeneous set of base learners and selects the learner that performed best in the previous window to make predictions for samples in the current window. Finally, online boosting performs boosting by weighting the base learners based on their prior performance and adjusting these weights over time \cite{Oza2005}. We will compare our AutoML strategies against several of these methods.

\subsubsection{Evaluation}
The most common procedures to evaluate online learning algorithms are holdout, prequential, and data-chunk evaluation. Since the data in a stream must be processed in their temporal order, cross validation is not applicable \cite{Gama2014}. Holdouts can be applied by requiring that only the earlier data is used for training and the later data for testing. In \textit{prequential evaluation} (or \textit{interleaved test-then-train}), newly arriving instances are first used to calculate the performance (test) and then for updating the model (train) in the next iteration. Accuracy changes incrementally as new instances and their labels arrive. Although this type of evaluation has been shown to yield lower accuracy scores than holdout on average \cite{Rijn2014}, it is very useful to evaluate models in case of concept drift.
%NEW UP
%NEW DOWN
A popular middle-way approach that combines holdout and prequential evaluation is \textit{data chunk evaluation}, which uses data chunks of size $s$ instead of individual instances to apply the test-then-train paradigm \cite{Bifet2009}. This approach doesn't punish algorithms for early mistakes and training time can be measured more consistently. 
%A critical aspect is determining the size of data chunks which is related to the drift type and computational cost.
We will therefore use data chunk evaluation in this paper.

%One of the disadvantages of individual prequential evaluation is algorithms being punished for early mistakes preventing them to reach their true accuracy. The advantages of the data chunk approach are its ability to measure training time, unlike classical prequential evaluation, and reducing the effect of accuracy obscuring \cite{Bifet2009}. One issue is to adequately determine the size of the data chunks, especially on data streams with concept drift, since it would have an effect on the learner's adaptation to the change. Small size chunks are more suitable for sudden drift as the change would be detected earlier, yet it may not perform as well as larger chunks for stable periods, and it increases the computational cost \cite{Brzezinski2010}.

\section{Adaptation Strategies}\label{sec:adaptation}
Since AutoML methods usually optimize an objective function against a static training set (see Eq. \ref{eq:1}), several modifications are required to apply them on evolving data. We call these modifications \textit{adaptation strategies} because they can be generally applied to adapt AutoML methods. Madrid et al. \cite{Madrid2019} introduced two such strategies, but then specifically for auto-sklearn. In \textit{global model replacement}, models are retrained on all prior data and recombined into ensembles based on meta-features when drift is detected. In \textit{model management} only the weights of the base models in autosklearn's final ensemble are updated based on their performance.
%NEW UP
\subsection{Strategy definitions}
\begin{figure}[t]
    \centering
    \includegraphics[trim=50 20 50 0, width=3in]{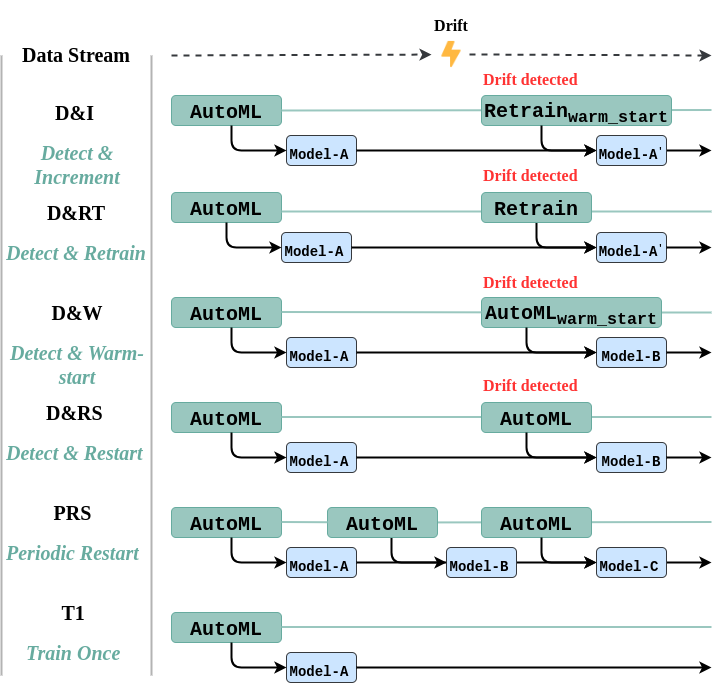}
    \caption{Adaptation strategies}
    \label{fig:strategies}
\end{figure}

In this paper, we propose and evaluate six different adaptation strategies, visualized in Fig. \ref{fig:strategies}. In each of the following strategies, the AutoML algorithm is run at least once at the beginning with the initial batch of data. For the forgetting mechanism we use a fixed length sliding window, where the last \textit{s} batches are used for training the learner and the earlier data is forgotten. The hyperparameter \textit{s} can be chosen depending on the application: lower values imply lower memory requirements and faster adaptation to concept drift, while higher values imply larger training sets and less frequent re-optimization of the pipelines.

%In addition to Madrid's global model replacement \cite{Madrid2019} and Bakirov's  no-adaptation \cite{Bakirov2018} mechanisms, we introduced a warm start approach and best model adaptation in case of drift with a sliding batch of new data. 

%NEW DOWN
\begin{description}
\item [D\&I Detect \& Increment] This strategy includes a drift detector that observes changes in pipeline performance and uses this to detect concept drift. AutoML is run on the first batch of data to build a first model (\texttt{Model-A}). We limit the AutoML search space to pipelines with incremental learning algorithms, i.e. learners that can continue to train on new data, such as random forests and gradient boosting. This restricts the AutoML methods, but it speeds up the retraining and we verified empirically that this still produces top-performing pipelines. %As long as there is no drift in the data, \texttt{Model-A} is used to predict all upcoming batches.%
\texttt{Model-A} is updated only if drift is detected, with the pipelines \textit{trained incrementally} with the latest \textit{s} (sliding window) batches. The configuration of the pipelines remains the same.  
%%% Commenting this out until we check how this is done.
%If the AutoML classifier is an ensemble, the base pipelines are each trained incrementally and their weights in the ensemble are kept to their original values. 
This strategy assumes that the initial pipeline configuration will remain useful and only the models need to be updated in case their performance dwindles because of concept drift. It tests whether keeping a learner memory of past data is beneficial. This can give a performance boost if the learner can also adapt to the new data. %Training incrementally is usually faster than training from scratch.
\vskip 0.07in
%It is also a comparison basis for evaluating the effect of rerunning AutoML throughout the data stream. 

\item [D\&RT Detect \& Retrain] This strategy is similar, but runs AutoML without restricting the search space and retrains the pipelines \textit{from scratch} on the latest \textit{s} batches when drift is detected. The original pipeline configurations and ensemble weights are kept. This strategy also assumes that the initially found pipeline configuration will remain useful and only the models need to be retrained in case of concept drift. It eliminates the need for rerunning the (expensive) AutoML techniques with every drift, but may be less useful if the drift is so large that a pipeline redesign is needed.
\vskip 0.07in
\item [D\&WS Detect \& Warm-start] After drift is detected, the AutoML technique is re-run to find a better pipeline configuration. Instead of starting from scratch, however, 
the AutoML is rerun with a 'warm start', i.e. starting the search from the best earlier evaluated pipeline configurations. This strategy assumes that the initial  pipeline configurations are no longer optimal after concept drift and should be re-tuned. Using a warm start mechanism can lead to faster convergence to good configurations, hence working better under small time budgets. However, in case of sudden drift the previously best pipelines may be misleading the search.
\vskip 0.07in
%The purpose is to keep a memory of the past data, and see its effect on the current performance. This strategy can also be used to decrease the maximum evaluation time resulting in some run time savings. A possible drawback is, in case of sudden drift, past data can harm the performance of algorithm by decreasing the adaptation time of the learner to the new settings. The hypothesis tested is whether information on past can be used to improve performance and worth keeping in memory. Since not all AutoML libraries include a warm start option from previous runs due to the batch setting with one run they are designed for, this strategy may require modifications to the libraries. 

\item [D\&RS Detect \& Restart]  After drift is detected, the AutoML technique is re-run from scratch, with a fixed time budget, to find a better pipeline configuration. This is a generalization of \textit{global model replacement} \cite{Madrid2019} to other AutoML systems and an extension of the hyperparameter optimization in \cite{veloso2018self,carnein2019towards} to full pipelines. This strategy also assumes that the pipelines need to be re-tuned after drift. Rerunning the AutoML from scratch is more expensive, but could result in significant performance improvements in case of significant drift.
\vskip 0.07in
\item [PRS Periodic Restart] Similar to 'Detect \& Restart', except that instead of using a drift detector, AutoML is restarted with each new batch, or some other regular interval. This strategy tests whether a drift detector is useful, and whether it is worth to retrain at certain intervals in spite of the significant computational cost.
\vskip 0.07in
\item [T1 Train once] This is a baseline strategy added for comparison purposes. AutoML is run once at the beginning and the resulting model is used to test each upcoming batch. This tests the innate capabilities of AutoML methods to cope with concept drift.
\end{description}

%Note that these strategies can be used whether the AutoML system returns the usual batch learning pipelines or online learning pipelines. However, we are not aware of any AutoML systems that designs online learning pipelines or could be easily adapted. Hence we could not explore this in this study, but recommend it as future work.

\subsection{Integration in AutoML Systems}
%Integrating these strategies requires several significant changes to existing AutoML systems. 
To evaluate the utility of these adaptation techniques, we have implemented them in Auto-sklearn, H2O and GAMA, representing AutoML approaches based on Bayesian optimization, random search, and evolution, respectively.

\textbf{Drift detection}. The first four adaptation strategies require a drift detector. We apply the Early Drift Detection Method (EDDM) \cite{BaenaGarcia2006} which is an improvement over the widely used DDM method \cite{Gama2004} that better detects gradual drift. EDDM assumes that concept drift occurs when the learner's misclassifications are spaced closer together over time. After every misclassified sample $i$, it computes the average distance between misclassifications $p_i$ and the standard deviation $s_i$ in the current sequence (or batch). It also records value $(p_i + 2s_i)$, the 95\% point of the distribution, and its peak value $(p_{max} + 2s_{max})$. Drift occurs when misclassifications are spaced more closely than usual:

\begin{equation}
(p_i + 2s_i) / (p_{max} + 2s_{max}) < \alpha
\end{equation}
The threshold $\alpha$ is suggested to be set to 0.95.
% When the threshold of 0.9 is passed, a warning is raised and the methods stores the samples in anticipation of concept drift.

\begin{table*}[t]
\centering
\caption{Implementation details for the AutoML method adaptations.}
\label{tab:2}       
\tabcolsep=1mm
\scalebox{0.85}{
\begin{tabular}{L{1.4cm}|L{7.1cm}|L{6.1cm}|L{5.4cm}}

\hline\noalign{\smallskip}
\textbf{Strategy} & \textbf{H2O Adaptation} & \textbf{Autosklearn Adaptation} & \textbf{GAMA Adaptation} \\

\hline

\noalign{\smallskip}\hline
\multirow{1}{*}Detect \& Increment
&
- Restrict the base models to GBM and RF, tune the hyperparameters with \textit{RandomGridSearch},
%for the space defined in \textit{h2o.automl} settings. 
and train a \textit{StackedEnsemble} on the best pipelines using the initial batch.

- When drift is detected, use the \textit{checkpoint} option to train the base models incrementally on new data with the same hyperparameters.

- Retrain the stacked ensemble with the updated models on the new data, with the same hyperparameters.
&
- Restrict the search space to GBM and RF, set option \textit{warm-start = True}, and run AutoML on the initial batch.

- When drift is detected, use the \textit{refit} function to retrain the pipelines retrieved from the current ensemble on the latest data. 
&
- Restrict the search space to GBM and RF, set option \textit{warm-start = True}, and run AutoML on the initial batch.

- When drift is detected, retrain the pipelines in the ensemble on the latest data by retraining the model retrieved from the previous AutoML run with \textit{automl.model.fit}. Use the same hyperparameters.
\\
\hline
\multirow{1}{*}
Detect \& Retrain 
&  
- Run AutoML on the initial batch.

- When drift is detected, retrain the best model (\textit{leader}). If this is a stacked ensemble, get the base models 
%(with \textit{get\_params()['base\_models']})
and retrain them on the new batches. 

- Retrain the stacked ensemble with the updated models on the new data.
&  
- Run AutoML on the initial batch.

- When drift is detected, use the \textit{refit} function to retrain the pipelines in the ensemble model from the original AutoML run on the latest data.
& 
- Run AutoML on the initial batch.

- When drift is detected, use the \textit{refit} function to retrain the pipelines in the ensemble model from the original AutoML run on the latest data with \textit{automl.model.fit}
\\

\hline
\multirow{1}{*}
Detect \& Warm-start &  
The best pipelines need to be retrieved from the previous \textit{RandomGridSearch} and re-evaluated in the next, before retraining the \textit{StackedEnsemble} on the best pipelines. Sadly, controlling the random grid search in this way is not currently supported in H2O. To approximate, we run a new random search and add the best new pipelines to a voting ensemble that also includes the previous best pipelines. 
& 
- To share the models between batches, run AutoML with options \textit{parallel usage with manual process spawning} and \textit{Ensemble\_size = 0}. Then, build an ensemble on the trained models.

- Warm-start the Bayesian optimization with the best previous pipelines. Keep 1/3 of the budget for building the ensemble (\textit{fit\_ensemble}). Use the default \textit{Ensemble\_size} and \textit{ensemble\_nbest}.
& 
Use the \textit{warm start} feature of the classifier fit function. This causes the evolutionary search algorithm to start from the best previous pipelines.
\\
 
\hline
\multirow{1}{*}
Detect \& Restart &
\multicolumn{3}{l}{\shortstack[l]{Use the drift detector after each tested batch. Retrain AutoML from scratch with a fixed time budget on the new data when drift is detected.}}   
\\

\hline
\multirow{1}{*}
Periodic Restart & 
\multicolumn{3}{l}{\shortstack[l]{Re-run the AutoML after every batch with a fixed time budget.}}
\\

\hline
\multirow{1}{*}
Train once &
\multicolumn{3}{l}{\shortstack[l]{No change. Only run the AutoML methods on the first batch.}}
\\

\hline
\end{tabular}}
\end{table*}

\textbf{Adapting AutoML techniques}
The remaining non-trivial adaptations required for each system are summarized in Table \ref{tab:2}. First of all, none of the AutoML methods come with a built-in test-then-train procedure. We therefore perform data chunk evaluation by dividing the data into \textit{n} batches and feeding them one by one in arriving order to the method for prediction/testing first, and training afterwards. 
%We simulate incremental training by retraining the pipelines after every batch with all batches seen since the beginning or since the last concept drift or periodic restart.
None of the methods cover online learning algorithms. For \textit{Detect \& Increment} we restricted the search space to gradient boosting (GBM) and random forest (RF) models, which are supported by all methods and allow incremental learning.

%Second, since all 3 AutoML systems build ensembles, retraining the pipelines requires deconstructing the ensemble to retrieve the base-learners and retrain them individually. 
%Finally, warm-starting the pipeline search also requires retrieving the best pipelines from the ensemble and warm-start the search with them. In H2O AutoML, we needed to reduce the search space to pipelines with gradient boosting and random forests only, since only these can be warm-started (using their checkpoint mechanism).

\section{Experimental Setup}\label{sec:experiments}
In this section we describe the data streams used to evaluate the adaptation mechanisms and the setup of the AutoML systems. To ensure reproducibility, all data streams are publicly available at OpenML \cite{Vanschoren2014} together with results of experiments with different algorithms.\footnote{Search \url{www.openml.org} for datasets tagged `concept\_drift`.} We also provide a github repository with our code and empirical results, including many plots that could not be fitted in this paper.\footnote{\url{https://github.com/openml/continual-automl}}

\subsection{Data streams}
We selected 4 well-known classification data streams with real-world concept drift, and generated 15 artificial ones with different drift characteristics. The latter are generated using the MOA framework \cite{Bifet2011}, and include gradual, sudden and mixed (gradual \& sudden) drift. Within each drift type, the drift magnitude is changed by changing the underlying distribution functions, and in some cases a certain amount of artificial noise is added. 

\begin{description}[font=$\bullet$~\normalfont\scshape]
\item [Airlines] is a time series on flight delays \cite{Gomes2017} with drift at daily and weekly intervals \cite{Webb2018}. It has \num{539383} instances and 9 numeric and categorical features.

\item [Electricity] is a time series on electricity pricing  \cite{Harries1999} that has been shown to contain different kinds of drift \cite{Webb2018}. It has \num{45312} instances, 7 features, and 2 classes. 
%Each instance corresponds to a period of 30 minutes with a total of 45,312 instances. Although it is not possible to make a certain conclusion about the drift existence and type, \cite{Webb2018} shows drift maps of sudden increase in covariate drift and fluctuating, relatively low levels of class probability drift. 

\item [IMDB] is a text data stream with data from the Internet Movie Database \cite{Read2012}. An often-used task in drift research is to predict whether a movie belongs to the "drama" genre. It has \num{120919} movie plots described by \num{1000} binary features.

\item [Vehicle] includes sensor data from a wireless sensor network for moving vehicle classification \cite{Duarte2004}. It has \num{98528} instances of \num{50} acoustic and \num{50} seismic features.

\item [Streaming Ensemble Algorithm (Sea)]  is a data generator based on four classification functions \cite{Street2001}.
%which uses the sum of 2 attributes out of 3 with a class threshold value that changes with the function used.
We created data streams of \num{500000} instances and \num{3} numerical features with concept drift. 
%In this study, both abrupt and mixed drift data streams are generated by SEA. 
Abrupt drift is created by changing the underlying classification function at instance \num{250000}. The drift window, $w$, is the number of instances through which the drift happens. The abrupt drift data streams were generated by switching between different functions with $w=1$ and introducing 10\% label noise.
%with an expected relative effect on the drift magnitude. The streams are introduced with .
We also generated mixed drift streams by additionally adding two gradual drifts (before and after the abrupt drift) with $w=\num{100000}$. The mixed drift data streams were generated with different magnitudes of sudden and gradual drift, without adding noise. 

\item [Rotating Hyperplane] is a stream generator that generates d-dimensional classification problems in which the prediction is defined by a rotating hyperplane \cite{Hulten2001}. By changing the orientation and position of the hyperplane over time in a smooth manner, we can introduce smooth concept drift. We created a \num{5} gradual drift data streams with different drift magnitudes within a window $w$ of \num{100000}.
% by increasing the factor of change after every instance.
The data streams contain \num{500000} instances with \num{10} features. 5\% noise is added by randomly changing the class labels. 
\end{description}

The data streams are divided into batches of equal size to simulate an online environment and given to the algorithms in arriving order, one batch at a time. We choose a batch size of \num{1000} to provide enough data for the AutoML algorithms and minimize the effect of accuracy fluctuations caused by small batches. For PRS, the batch size is larger ($\sim$\num{20000} instances) since it does not include a drift detector and requires retraining with every batch, which is much more expensive. We apply data chunk evaluation to evaluate the adapted AutoML systems. For training, a  sliding window with a fixed length of 3 batches is chosen for each dataset.

\subsection{Algorithms}
% add more information on default and changed parameters based on section 3
We evaluate the adaptations of Autosklearn, GAMA and H2O. As baselines, we added Oza Bagging \cite{Bifet2009} and Blast \cite{vanRijn2016}. Both are state-of-the-art online ensemble methods specifically designed for data streams with concept drift. We add gradient boosting as a baseline for incremental learning. We explain how each algorithm is configured. Hyperparameters that are not mentioned are used with their default setting. 

\begin{description}[font=$\bullet$~\normalfont\scshape]

\item [Auto-sklearn] The time budget for each AutoML run is the default 1 hour. The memory limit is increased to \SI{12}{GB} to avoid memory errors. 

\item [GAMA] The time budget for each AutoML run is set to 1 hour. The option to build an ensemble out of the trained machine learning pipelines is activated.

\item [H2O]  The time budget for each AutoML run is set to 1 hour. All settings are kept at their defaults, including the default stacker, a Generalized Linear Model (GLM).

\item [Oza Bagging] from the scikit-multiflow library \cite{Oza2001}. The window size is set to the batch size, \num{1000}, and the pretrain size is also set to \num{1000}. This way, training and prediction happens in the same way as for the  AutoML methods, allowing a fairer comparison.

\item [Blast] from the MOA library \cite{vanRijn2016}. The hyperparameters used are the defaults, and the window size is set to \num{1000}, similar to the AutoML implementations. 

\item [GBM] Gradient boosting from the scikit-learn library \cite{friedman2001}. The hyperparameters are set to their defaults and the window size is \num{1000}. A drift detector is used to retrain the model when drift is detected. 
%NEW END
\end{description}

Different drift detection algorithms were tested in multiple data streams prior to the selection. EDDM stands out as a robust and fast choice for different types of drifts, and hence used in all experiments.
%NEW END

\section{Results}\label{sec:results}
We evaluate all algorithms on all data streams, and analyze the effects of drift characteristics, adaptation strategies and AutoML approaches. 

\subsection{Synthetic data streams}

\begin{figure*}[!htbp]
\centering
\subfigure[][]{\includegraphics[trim=200 30 120 0,width=3.05in]{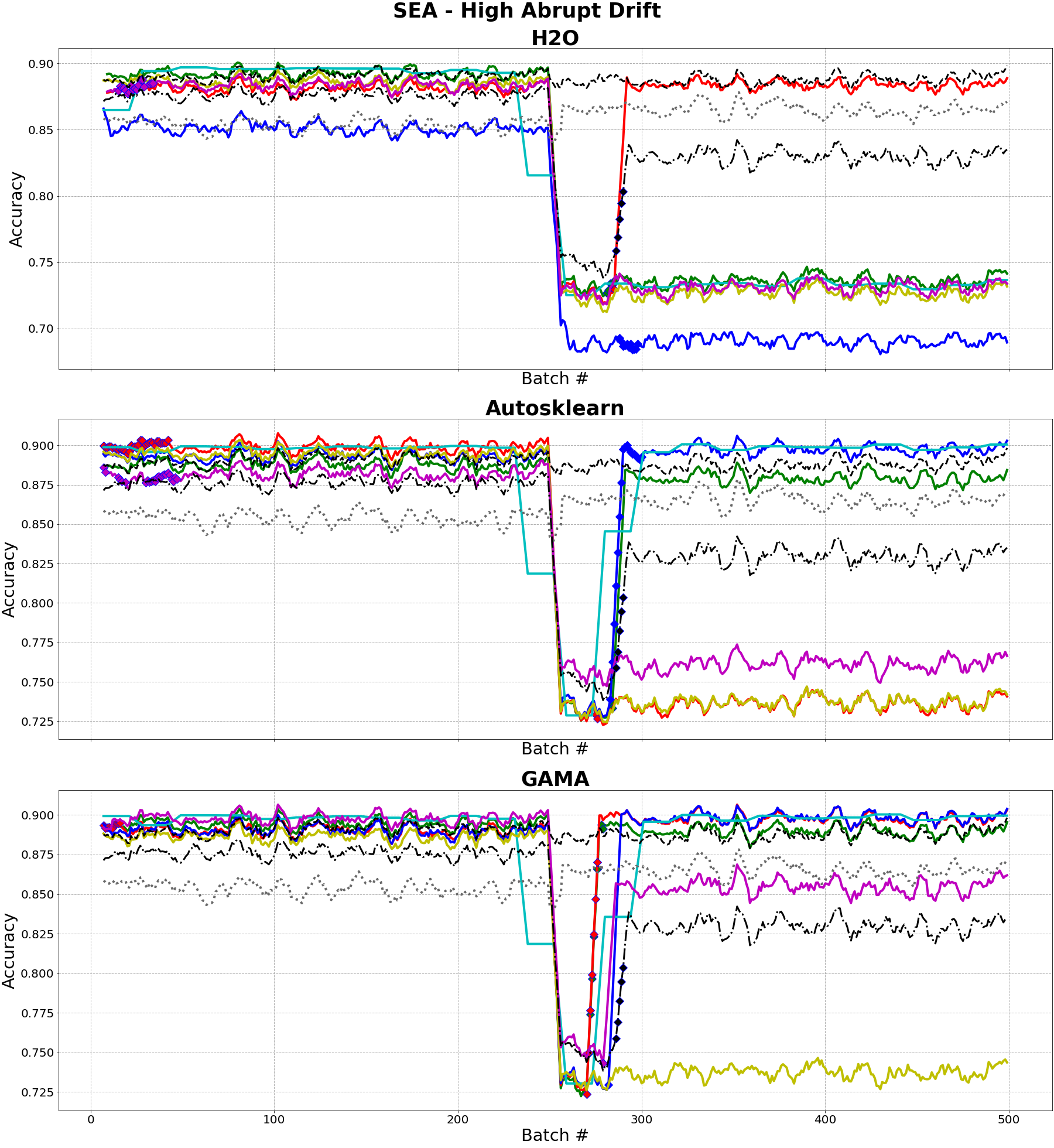}%
%\label{fig:2}-a
}
\hspace{1.5cm}
\vspace{0.3cm}
\subfigure[][]{\includegraphics[trim=200 30 120 0,width=3.05in]{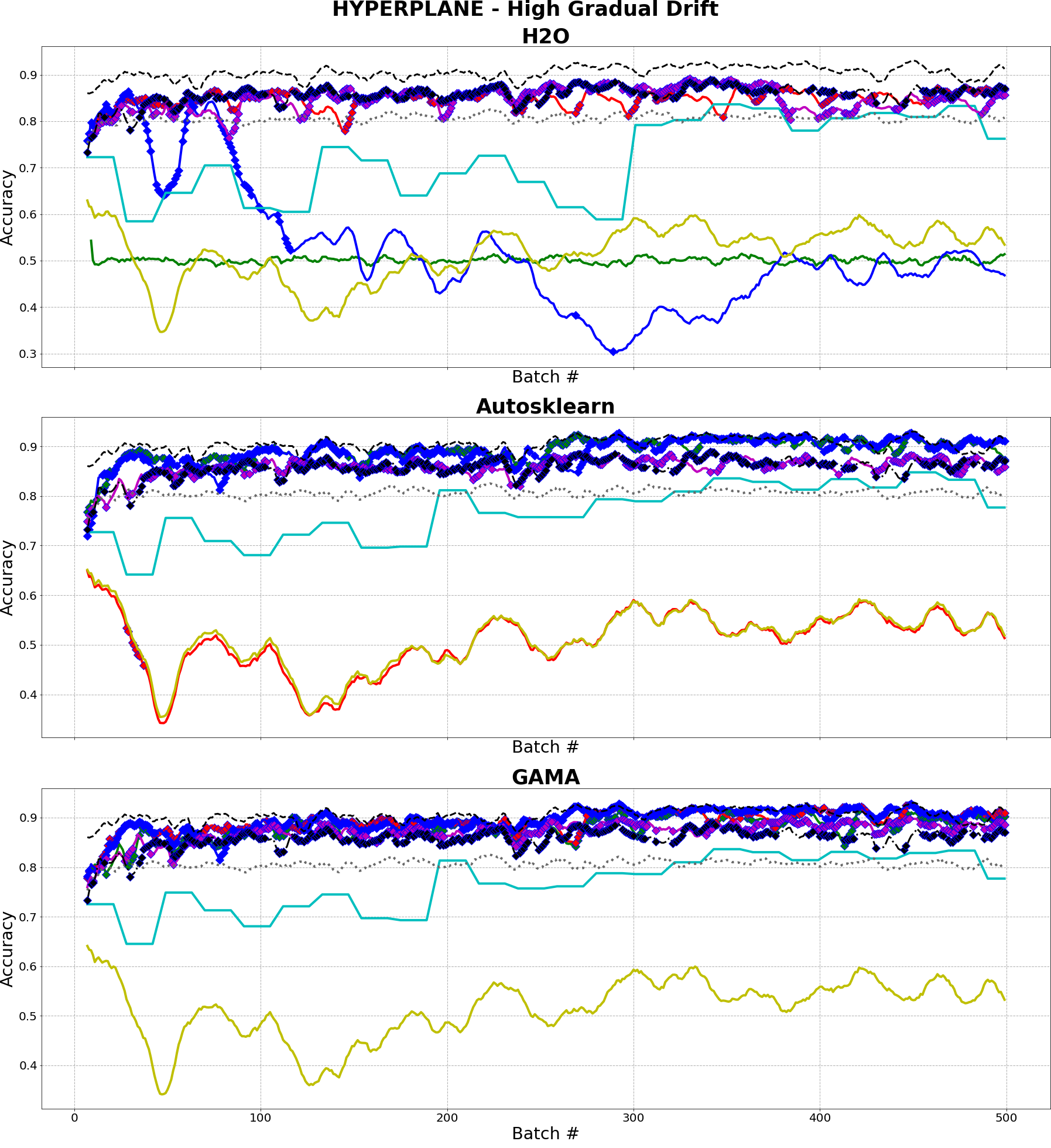}%
%\label{fig:2}-b
}
\hspace{1.5cm}
\vspace{0.3cm}
\subfigure[][]{\includegraphics[trim=200 30 120 0,width=3.05in]{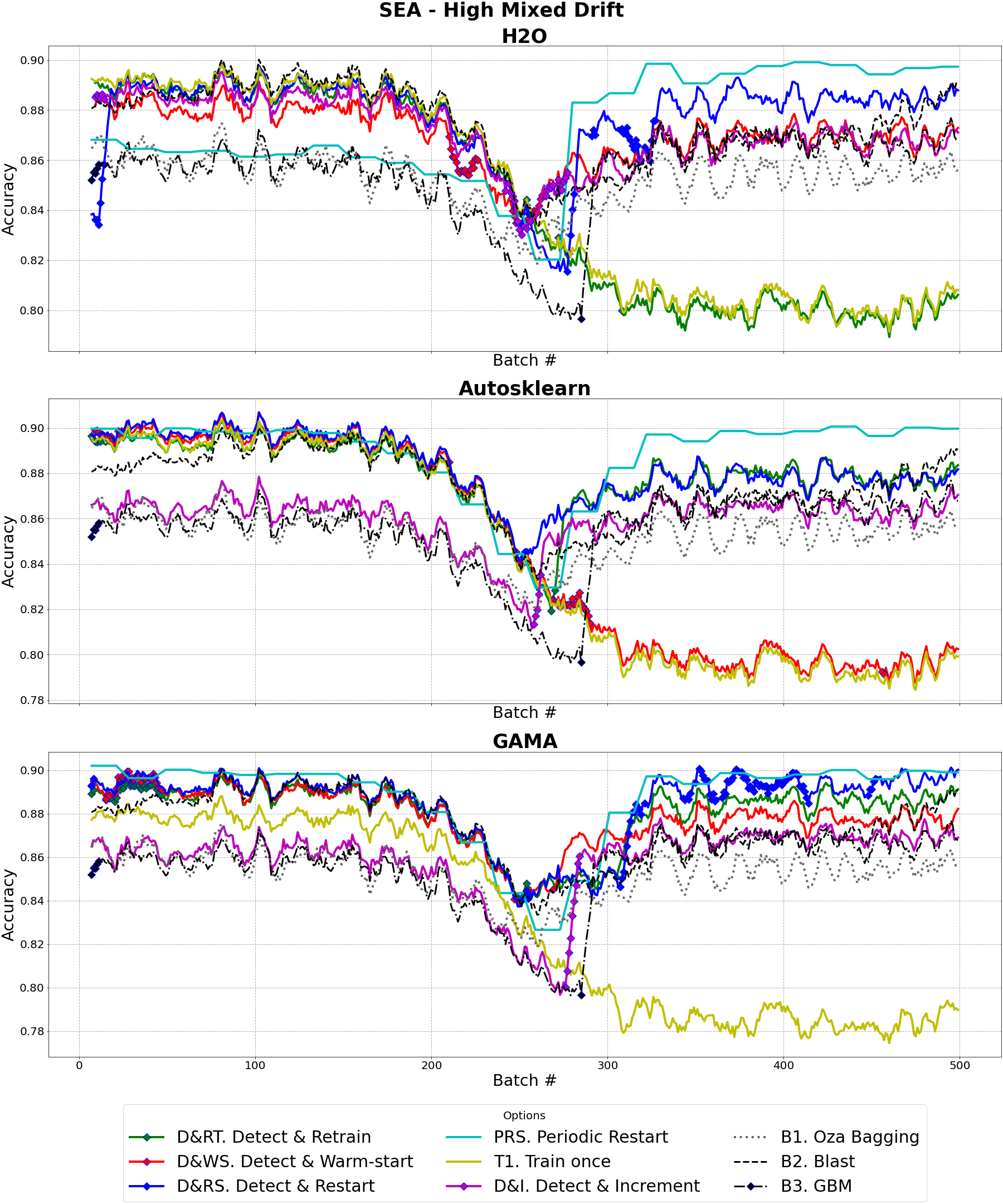}%
%\label{fig:2}-c
}
\hspace{1.5cm}
\vspace{0.3cm}
\subfigure[][]{\includegraphics[trim=200 30 120 0,width=3.07in]{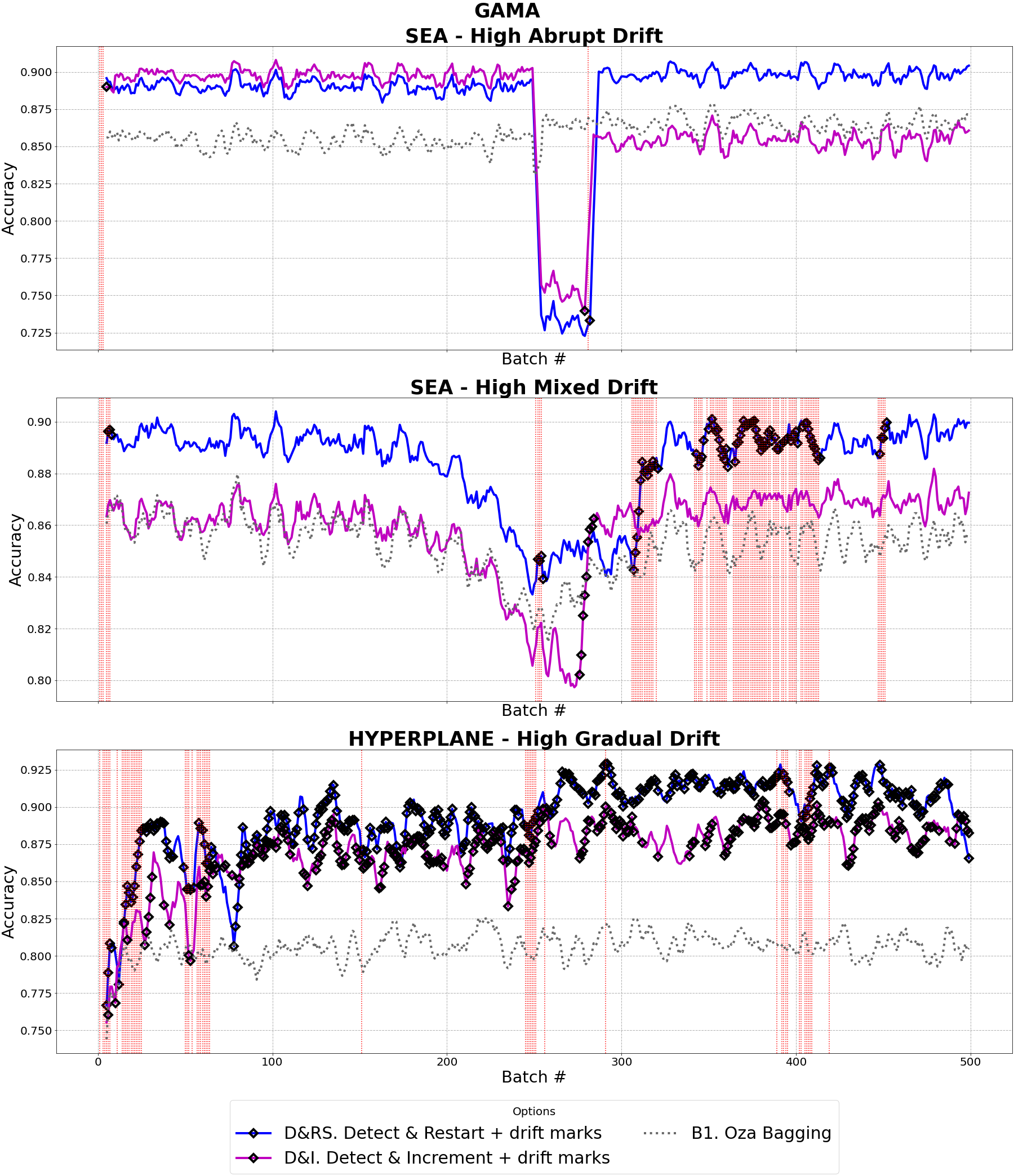}%
%\label{fig:2}-d
}
\vspace{-0.8cm}
\caption{Accuracy across batches for artificial data streams:
(a) SEA - High abrupt drift;
(b) HYPERPLANE - High gradual drift;
(c) SEA - High mixed drift, and 
(d) All three artificial data streams with drift marks and pipeline change points (red)
}
\label{fig:2}
\end{figure*}

We first analyze the effect of two important drift characteristics: drift type and drift magnitude. Fig. \ref{fig:2}-a to \ref{fig:2}-c demonstrate the results of artificial data streams with gradual, abrupt and mixed drift, respectively. Each subgraph shows the results of one AutoML library, and each series shows the accuracy for a specific adaptation strategy after every batch. The markers on the lines show when drift was detected. PRS plots are smoother since the batch size was increased for efficiency. These figures show the results for the highest drift magnitude (for clearer analysis). Results for different levels of drift magnitude will be shown in Fig. \ref{fig:magnitude}. %Also see Fig. \ref{fig:mean}: it contains the same results, but averaged over 20 batches (hence less noisy).

\subsubsection{Effect of Drift Type: High Abrupt Drift}
As shown in Fig. \ref{fig:2}-a, high abrupt drift severely affects all AutoML methods, and only some adaptation strategies help some of the AutoML methods to recover. GAMA is the quickest to recover after abrupt drift occurs when using warm-started AutoML (D\&WS) or when retraining the previously best pipelines after the drift is detected (D\&RT). Re-optimizing the pipelines from scratch after drift detection (D\&RS) works for both GAMA and Autosklearn, but takes a bit longer to recover. H2O only recovers with warm-starting (D\&WS). Periodically re-optimizing the pipelines from scratch without drift detection (PRS) also works for GAMA and AutoSklearn, but is of course more expensive. Blast is relatively unaffected by the abrupt drift and trails only slightly behind the recovered AutoML strategies. Oza Bagging is also unaffected by drift but performs significantly worse. GBM recovers after a delay but never regains its prior performance.

There are a few strategies that fail to recover at all. The default strategy of not updating the pipeline (T1) completely fails after sudden drift occurs. Incrementally updating the best pipeline (D\&I) fails with Autosklearn and H2O, and never fully recovers with GAMA either. This indicates that after sudden drift, pipelines need to be either retrained or re-optimized. Interestingly, warm-starting (D\&WS) doesn't help Autosklearn to recover, likely because the previously best pipelines are misleading its Bayesian Optimization approach. GAMA uses an evolutionary approach which does manage to evolve the previous pipelines into better ones. Surprisingly, for H2O also retraining (D\&RT) and re-optimizing the pipelines (D\&RS) fail to recover. This could be because the stacking procedure, a generalized linear model (GLM) by default, overfits on the predictions of the updated pipelines. This could potentially be solved with a more adaptive stacking method (e.g. GBM), larger batches, or maybe by feeding in the data points as well (cascade stacking). We will analyze this in more detail in Section \ref{sec:detection}.

%\textbf{Batch size}. The last strategy PRS (Periodic Restart), %without a drift detector, retrains the AutoML system from scratch, after adding each new batch to the sliding window. One disadvantage with this strategy is the
%requires a lot more training time and possibly leads to more volatile behavior, which may be a disadvantage in certain data streams. This is why we increased the size of each batch and consequently decreased the number of retrainings. For abrupt and mixed drifts (Fig. \ref{fig:2}-a and \ref{fig:2}-c respectively), its performance is quite similar to the options that include a drift detector, even slightly better and recovering faster in mixed drift after the sudden drift point. For gradual drift (Fig. \ref{fig:2}-b), on the other hand, PRS fails to reach the level of drift detector options in all libraries for almost every batch. This might be due to overfitting to temporary conditions occurring in continuous drift whereas drift detectors only trigger when the drift if significant enough. With the additional computational burden of retraining with every batch and lower accuracy, this strategy seems less useful in online learning scenarios.
%NEW END

\subsubsection{Effect of Drift Type: High Gradual Drift}
As shown in Fig. \ref{fig:2}-b, D\&RT and D\&RS perform well under high gradual drift in both GAMA and Autosklearn, but not for H2O (again likely due to the GLM stacker overfitting). D\&I works quite well for all AutoML techniques. D\&WS works well for GAMA, but not for Autosklearn (again likely due to the Bayesian surrogate model being misled). Note that Blast is on par with the best options in GAMA and Autosklearn, and better than any option in H2O. OzaBagging is dominated by several AutoML adaptations. GBM performs better but still slightly worse than the best AutoML adaptations. The T1 baseline again performs poorly. PRS is also suboptimal, likely because the periodic restarts are not frequent enough and continually lag the high gradual drift.

\subsubsection{Effect of Drift Type: High Mixed Drift}
Fig. \ref{fig:2}-c shows that in case of high mixed drift, restarting the AutoML on new data (D\&RS) performs best overall. Compared to the abrupt drift data stream (Fig. \ref{fig:2}-a, AutoML methods handle the sudden drift better and recover faster. This is likely because the gradual drift prior to and after the sudden drift alarms the drift detectors periodically and causes more frequent retraining. Oza Bagging and Blast handle the drift well, yet overall fail to match the performance of the best AutoML strategies (D\&RS and D\&RT), especially after the sudden drift point. GBM performs similarly to Oza Bagging, but with a better recovery after the sudden drift. As in Fig. \ref{fig:2}-b, D\&RT doesn't work well for H2O and D\&WS doesn't work well for Autosklearn, likely for the same reasons. PRS recovers exceptionally well after the sudden drift, showing the advantages of re-running AutoML when significant drift occurs.
%NEW UP 

%Overall, Figures \ref{fig:2}-a-\ref{fig:2}-c show clear differences in accuracy changes over time as a result of the drift type, independent of the applied adaptation strategy and AutoML library. We can also observe the existence of more robust strategies that perform better than others with each drift type. 

%NEW
\subsubsection{Pipeline analysis}
In order to gain more insight into the underlying reasons behind these performance differences, pipelines at consecutive training points are compared. Fig. \ref{fig:2}-d indicates when pipelines change (vertical red lines), and when drift is detected (black line marks) for each of the three artificial data streams and strategy D\&RT. The accuracy plots of strategies D\&RT and D\&I are compared in addition to the Oza Bagging baseline. Retraining without re-optimizing the pipeline (D\&I) can outperform the baseline but is generally worse than re-optimizing the pipeline (D\&RT). This difference is more clear after the abrupt drift points. Therefore, while an initial AutoML optimization improves the performance compared to a baseline learner, re-optimizing the pipeline throughout the data stream can lead to different pipeline settings and clear advantages over static pipelines. The pipeline change points indicate that D\&RT indeed finds new pipelines regularly, especially under gradual or mixed drift.
%NEW END

\subsubsection{Effect of drift magnitude}
%Another important drift characteristic that is expected to have an effect on the learner performance and adaptation strategy (See section \ref{sec:problem}) is drift magnitude. The experiments are conducted to see whether level of drift magnitude would effect the best performing adaptation strategy for the same drift type and how much it would affect the performance of AutoML learners. 

\begin{figure}[t]
    \centering
    \includegraphics[trim=150 20 120 0,width=3.05in]{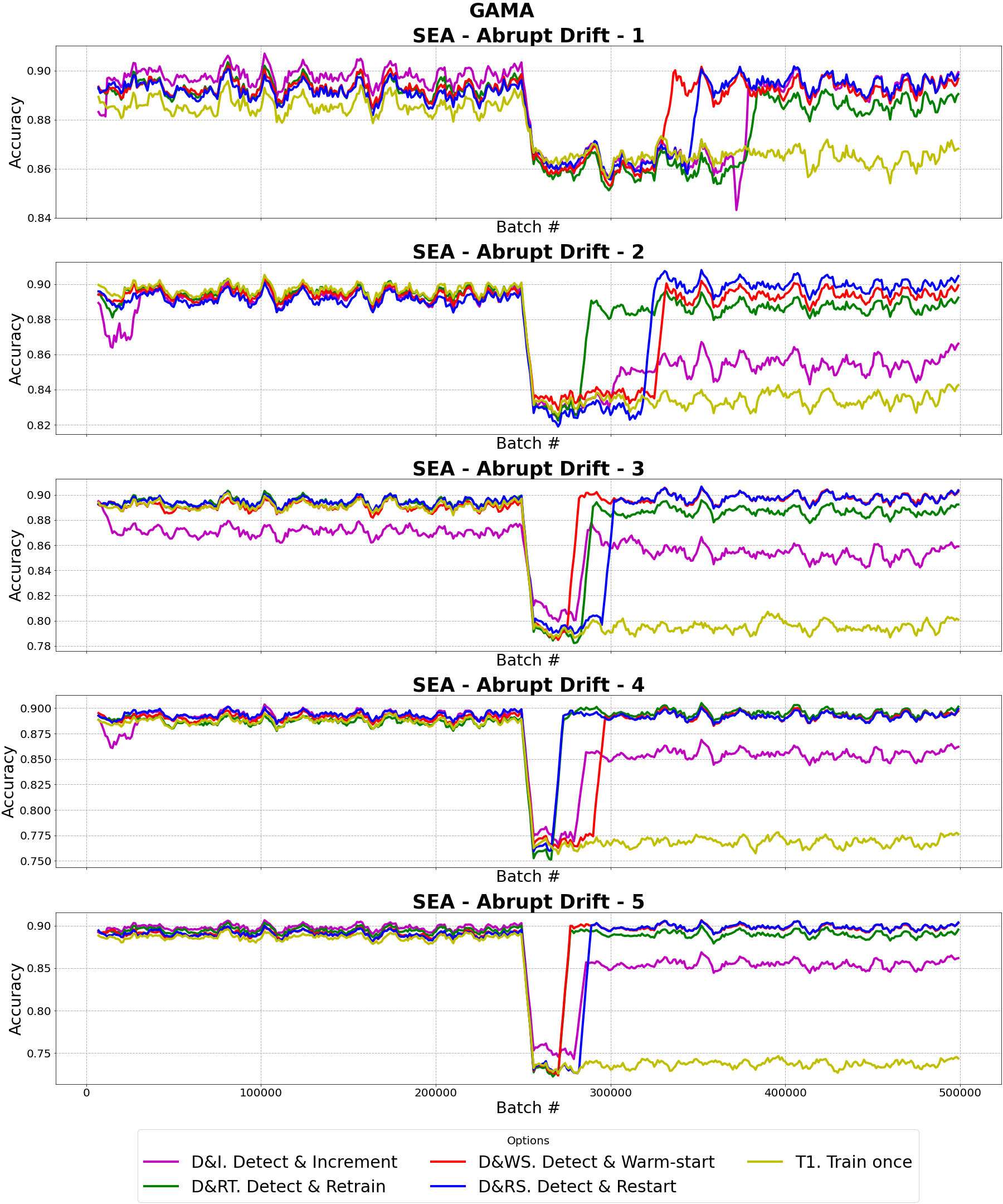}
    \caption{Effect of increasing abrupt drift magnitudes (1 is lowest magnitude) on different strategies, for GAMA.}
    \label{fig:magnitude}
\end{figure}

Fig. \ref{fig:magnitude} shows the results of GAMA library for increasing levels of \textit{drift magnitude} on the SEA abrupt drift data stream. As the drift magnitude increases, the performance drop also increases, but the overall recovery period \textit{decreases}. This is most likely due to the early triggering of the drift detector in high magnitude cases. The best adaptation strategy changes with the drift magnitude, as previously suggested in \cite{Madrid2019}. Retraining the models (D\&RT), however, generally recovers faster than restarting the AutoML (D\&RS). Warm-starting (D\&WS) sometimes recovers faster, but not always. The latter may depend on exactly \textit{how} the data drifts. If the concept underlying the data is still somewhat similar after the drift, i.e., it can be modelled with similar pipeline configurations, then warm-starting will speed up recovery, but if not, warm-starting could also be unhelpful.

%For lower levels of magnitude, the accuracy levels of different strategies are closer to each other since drift detectors are not as effective as with higher magnitude, more obvious drifts (hence eliminating the differences). As the magnitude increases, the strategy that recovers fastest after the drift changes between D\&RT and 2 reaching a very similar level at the end with D\&RS behind them. Overall, the results confirm the importance of the drift magnitude as well as drift type in design of AutoML adaptation for online data. 

\begin{figure*}[!htbp]
\centering
\subfigure[]{\includegraphics[trim=150 20 120 0,width=3in, keepaspectratio,valign=t]{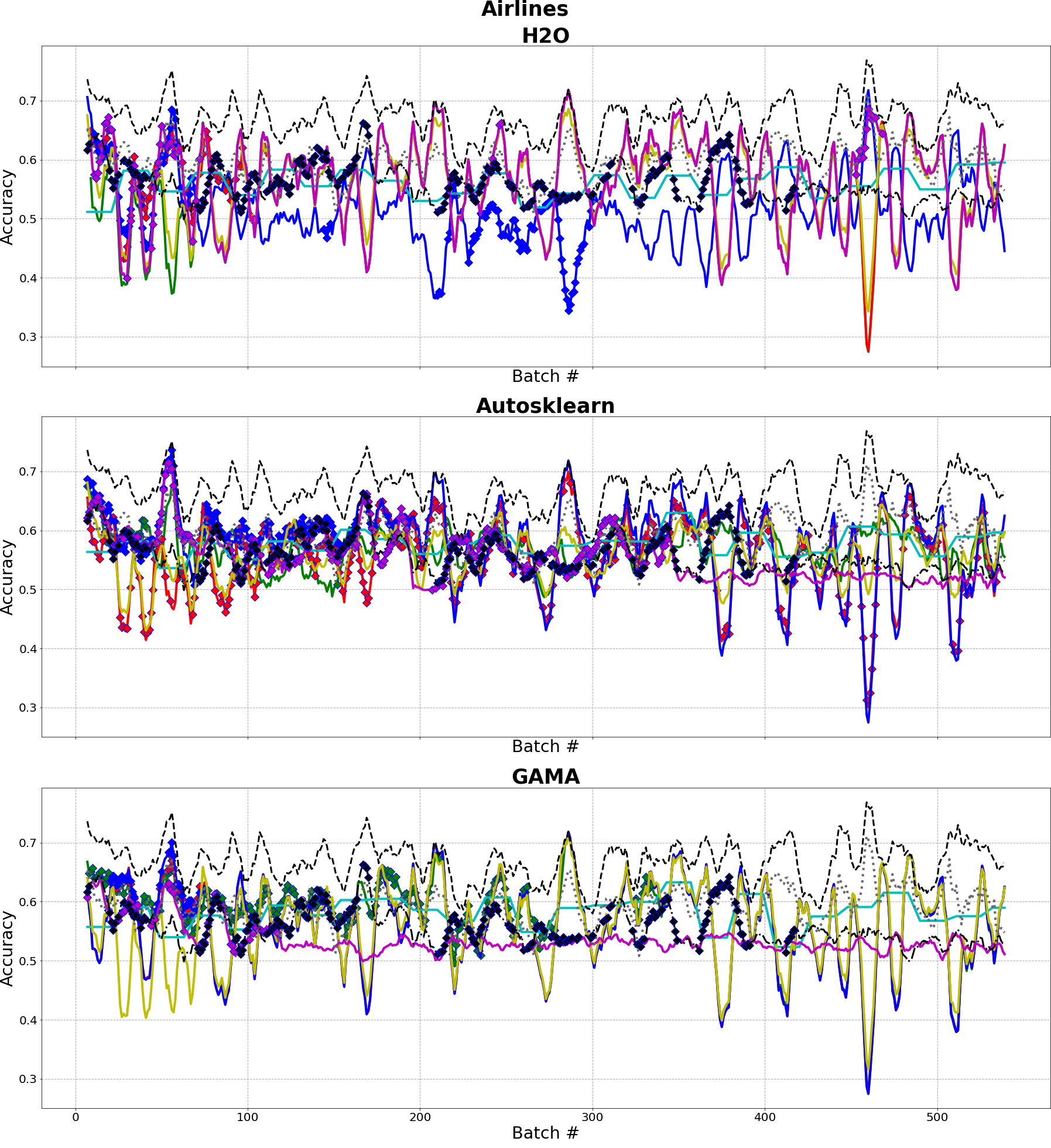}%
%\label{fig:3}-a
}
\hspace{1.5cm}
\subfigure[]{\includegraphics[trim=150 20 120 0,width=3in, keepaspectratio,valign=t]{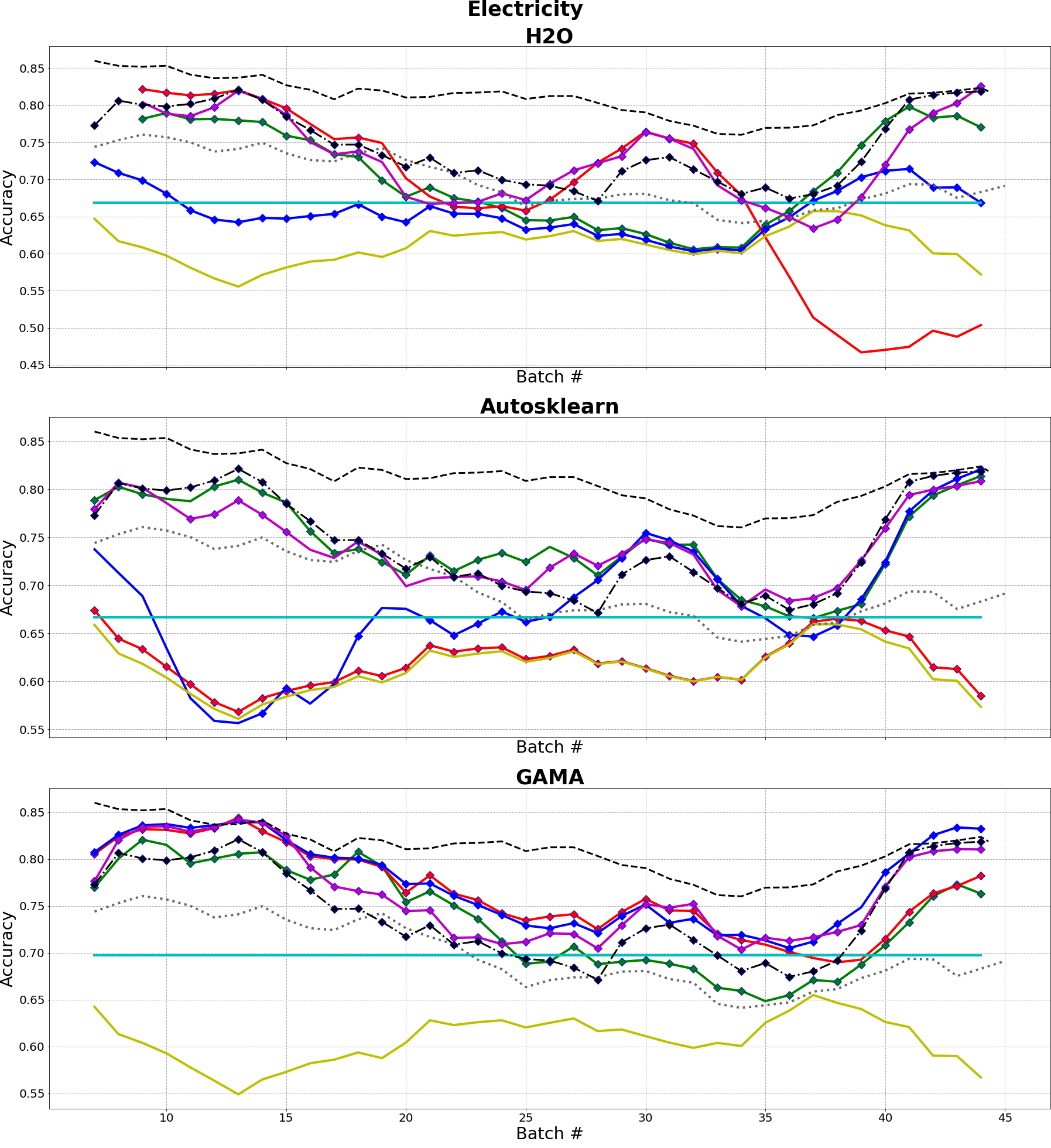}%
%\label{fig:3}-b
}
\hspace{1.5cm}
\subfigure[]{\includegraphics[trim=150 20 120 0,width=3in, keepaspectratio,valign=t]{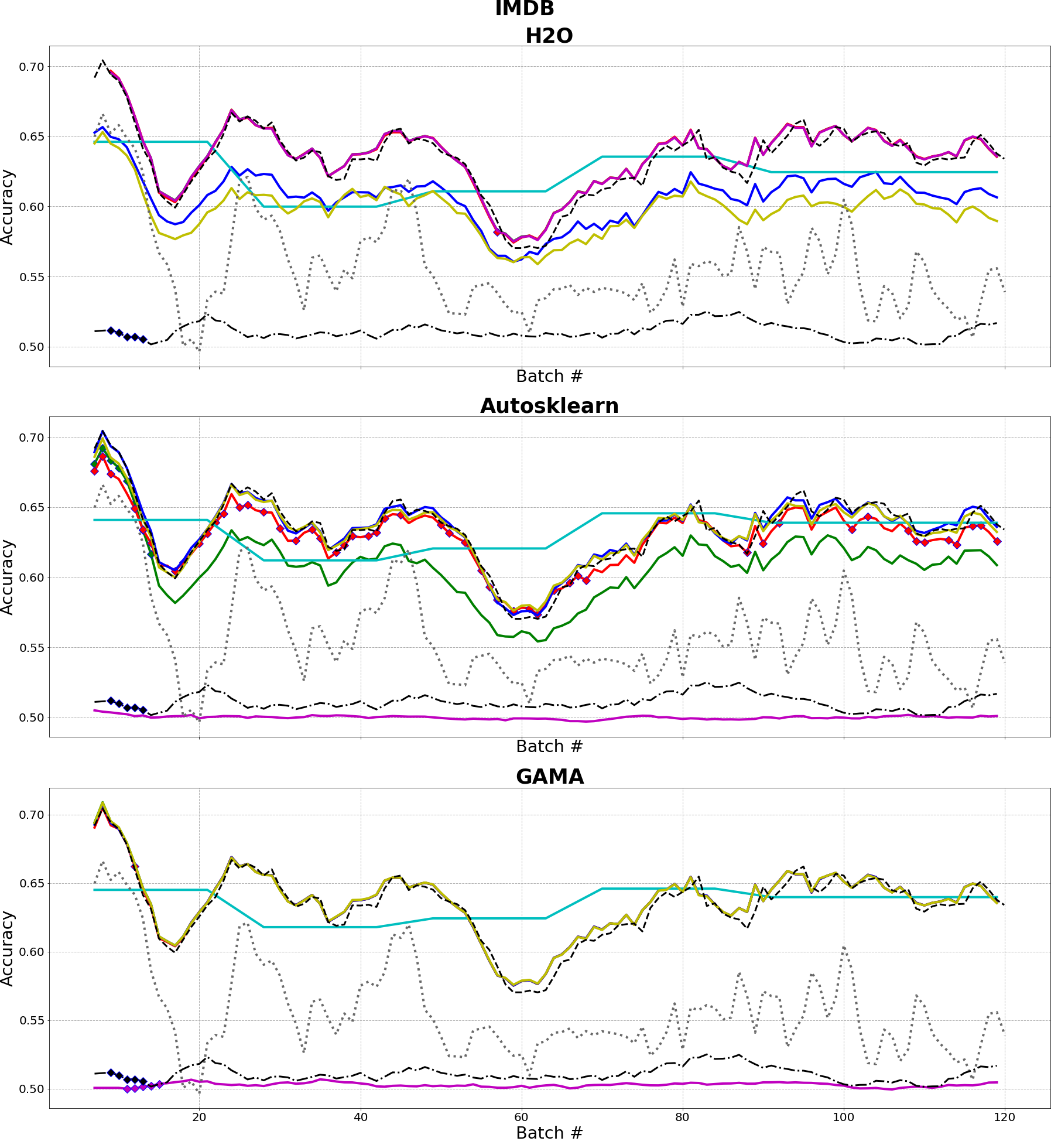}%
%\label{fig:3}-c
}
\hspace{1.5cm}
\subfigure[]{\includegraphics[trim=150 20 120 0,width=3in, keepaspectratio,valign=t]{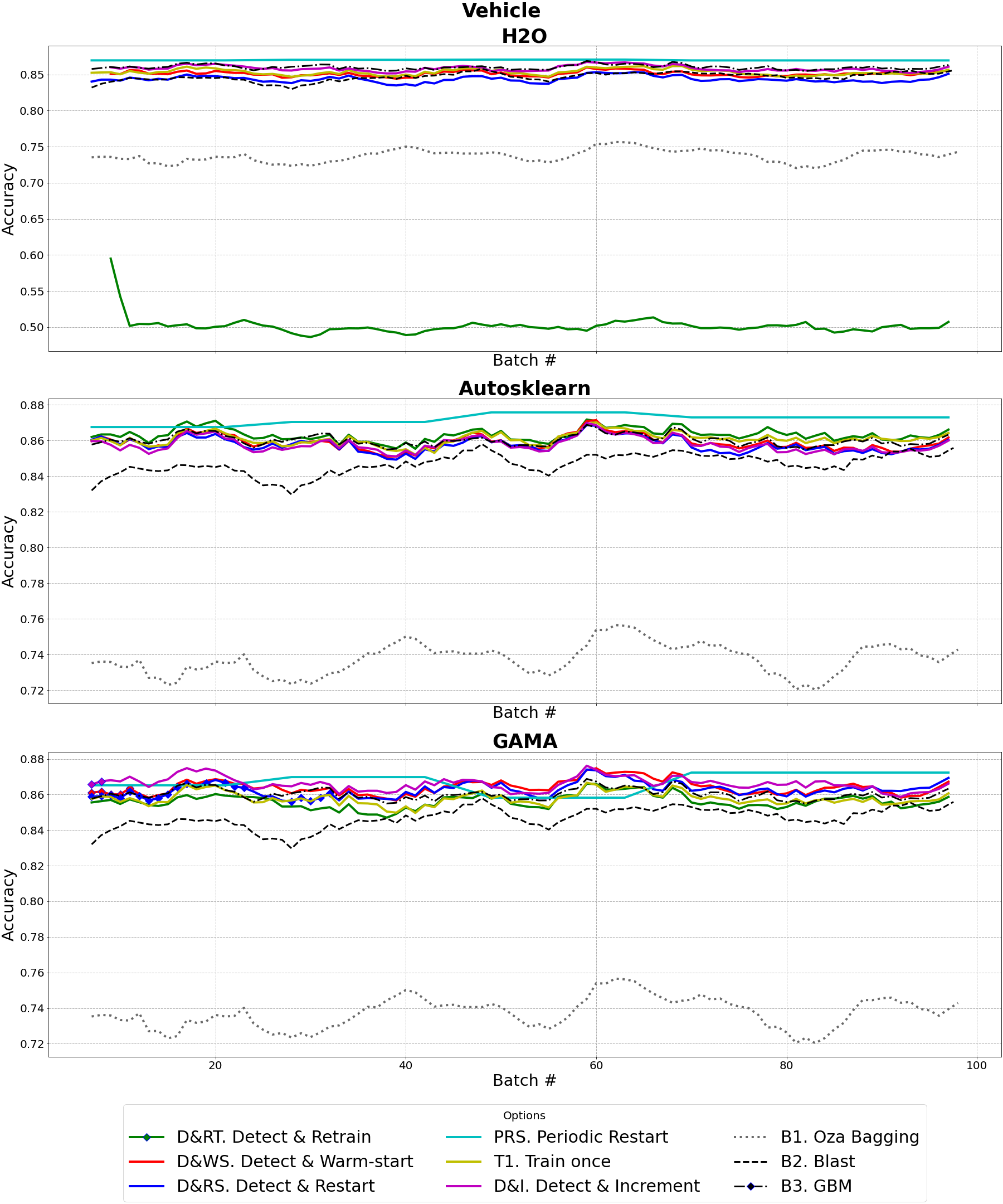}%
%\label{fig:3}-d
}
\caption{Accuracy across batches for the real data streams:
(a) Airlines;
(b) Electricity;
(c) IMDB; and,
(d) Vehicle}%
\label{fig:3}
\end{figure*}

\subsection{Real Data Streams}
The results on the real-world data streams are shown in Fig. \ref{fig:3}-a to Fig. \ref{fig:3}-d. The Airlines data (Fig. \ref{fig:3}-a) has particularly frequent drift. D\&RT and D\&I clearly perform worse here, indicating that sticking to the originally optimized pipeline is not ideal. PRS doesn't adapt quickly enough: restarting AutoML with every batch could be better but is also prohibitively expensive. The other adaptation strategies behave roughly the same, especially for GAMA where they overlap almost completely. D\&WS or D\&RS work generally well. H2O is again the exception: D\&RS does not recover well from drift (as observed earlier) and the other strategies fluctuate significantly. Blast works particularly well here. Oza Bagging is often outperformed by the best AutoML method but also fluctuates less. On IMDB data (Fig. \ref{fig:3}-c) we can make similar observations, although the drift is more gradual here, allowing PRS to find better pipelines. Blast is roughly on par with the AutoML strategies, while Oza Bagging and GBM are clearly worse. 

Electricity (Fig. \ref{fig:3}-b) is much smaller and has different kinds of drift, and hence we can observe more variation in the behavior of the different strategies and AutoML methods. Drift is detected with almost every batch. With GAMA, most strategies behave similarly as before. D\&RT and D\&I perform worse and D\&WS or D\&RS work generally well. AutoSKlearn and H2O have more difficulty with the smaller dataset, and re-optimizing the pipelines (D\&WS or D\&RS) often leads to worse results. PRS is a flat line because there aren't enough batches to warrant rerunning the AutoML techniques from scratch. Blast dominates the other methods here, GBM is mostly on par, Oza Bagging is much worse. The Electricity dataset is known to be heavily autocorrelated, which benefits Blast. On Vehicle, which is larger and had more gradual drift, both Blast and Oza Bagging are outperformed by the AutoML techniques and GBM. Note that few drifts are detected here, and hence the scores of the different strategies are very close. PRS works very well here, while H2O with D\&RT performs badly, as we've seen earlier with high gradual drift.
%NEW UP

%due to the custom adjustments done to the library creating a longer memory than required by a dynamic environment that emerges with certain drift characteristics. 

%\begin{figure}[]
%\centering
%\includegraphics[trim=150 20 120 %0,width=2.8in]{Mean_plots_data10.png}%
%\caption{Mean accuracy across batches under high abrupt drift}
%\label{fig:mean}
%\end{figure}

%\begin{figure}[]
%\centering
%\includegraphics[trim=150 20 120 0,width=2.8in]{Mean_plots_data16.png}%
%\caption{Mean accuracy across batches for HYPERPLANE - High gradual drift}%
%\label{fig:5}
%\end{figure}

\subsection{Comparison of AutoML systems}

In all, we found that the ideal adaptation strategy depends on the AutoML system and the characteristics of the data stream. On the real-world data streams (Fig \ref{fig:3}), each AutoML system performs quite similarly when selecting the best strategy, but the best strategy differs between them. On large datasets, re-optimizing pipelines (e.g. D\&WS or D\&RS) works well for GAMA and Autosklearn. The cheaper retraining without re-optimization (D\&RT) also works well on some datasets if regularly triggered to retrain. For H2O, D\&RS is met with overfitting issues, but incremental learning (D\&I) performs surprisingly well. On smaller datasets (e.g. Electricity), GAMA continues to re-optimize the pipelines well (D\&WS and D\&RS), while Autosklearn and H2O get better results by retraining the existing pipelines (D\&RT and D\&I). For data streams with abrupt concept drift, fast recovery is key. Restarting AutoML when abrupt drift is detected (D\&RS) is a safe bet for GAMA and Autosklearn. GAMA can recover even faster when warm-starting the evolutionary search (D\&WS), but this doesn't work with Autosklearn's Bayesian Optimization approach. This difference in recovery speed is bigger when the drift magnitude gets lower (Fig. \ref{fig:magnitude}). Similar behavior is observed for mixed drift (Fig. \ref{fig:2}-c), whereas GAMA and Autosklearn perform quite similarly on gradual drift. H2O did not recover with D\&RS in our experiments, which will be discussed in more detail next.

\begin{figure}[!t]
\centering
\subfigure[][]{\includegraphics[trim=150 20 120 0,width=2.8in]{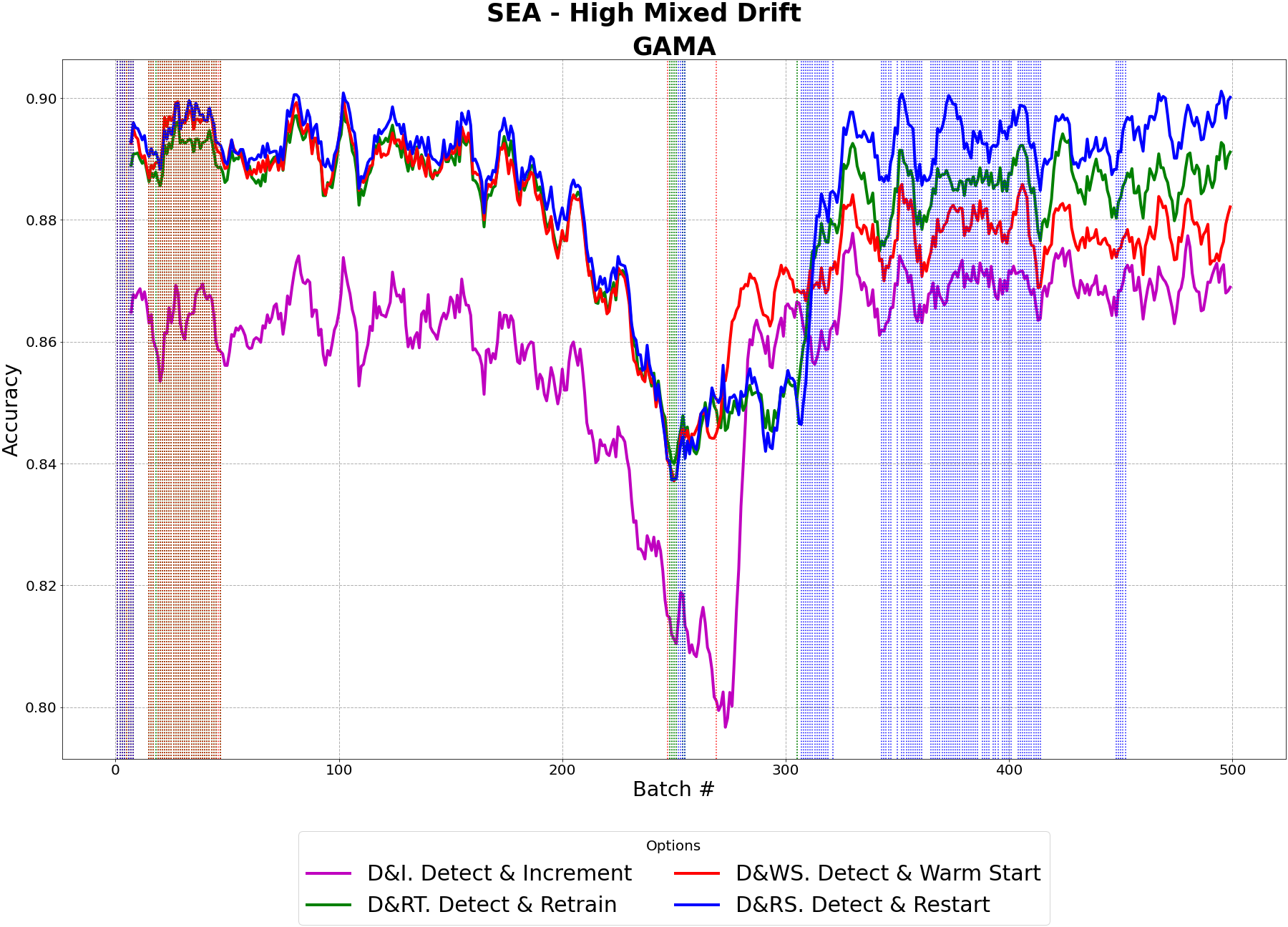}}
\hspace{1.5cm}
\vspace{0.3cm}
\subfigure[][]{\includegraphics[trim=150 20 120 0,width=2.8in]{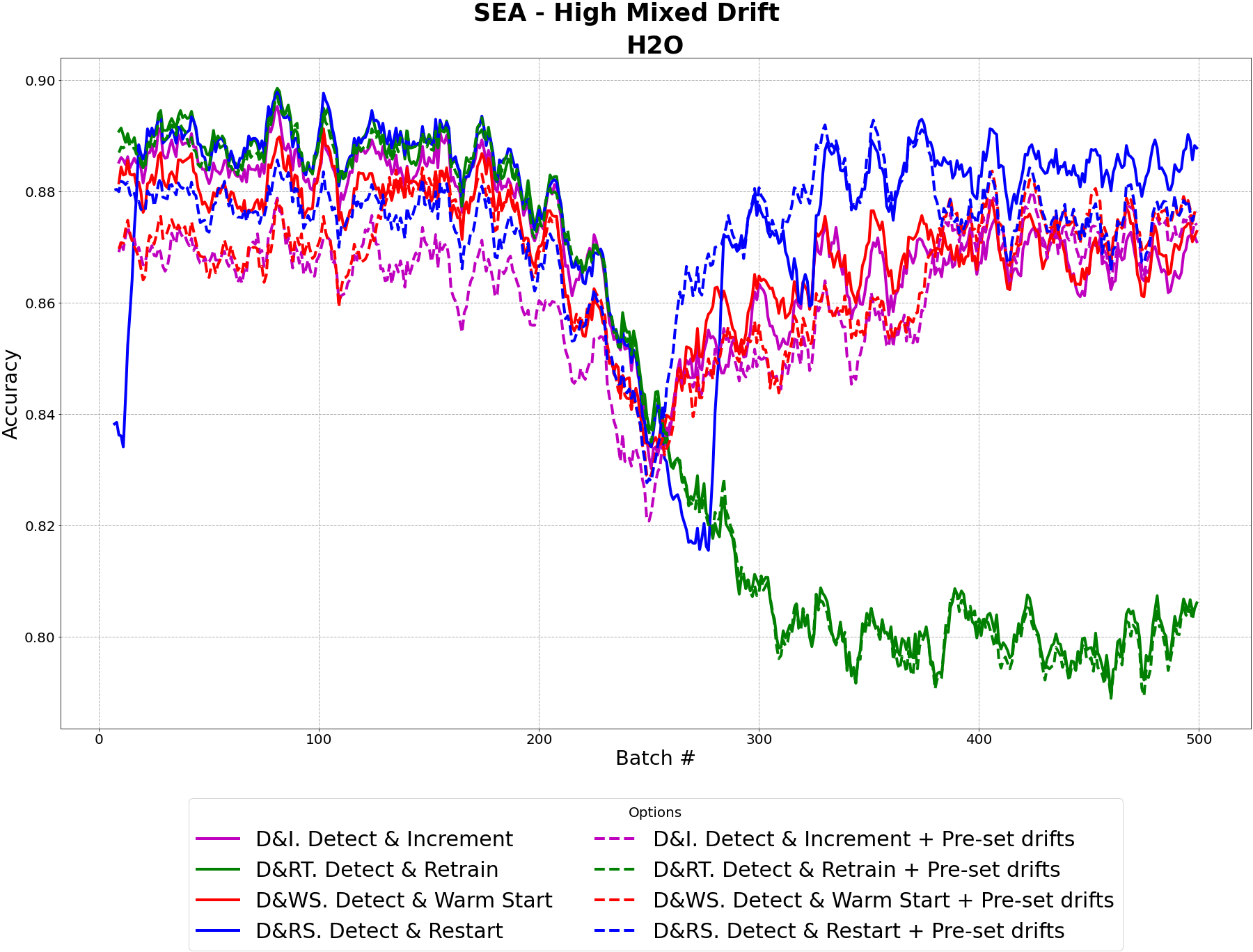}}
\caption{Accuracy across batches on SEA - High mixed data stream for:
(a) GAMA with detected drift points (vertical lines); and
(b) H2O with detected and pre-set drifts}%
\label{fig:drifts}
\end{figure}

%In the experiments with different magnitudes of abrupt drift, H2O is the most affected library with no options recovering for the lower levels of drift. 

\subsection{Interplay with drift detection}
\label{sec:detection}
To understand the interplay between the adaptation strategy and the drift detector, Fig. \ref{fig:drifts}-a shows the drift points (as vertical lines) for each adaptation strategy with GAMA on a mixed drift data stream. Re-optimizing the pipelines (D\&RS) leads to more significant performance differences, which trigger the drift detector more often, in turn leading to more re-optimization. Warm-starting (D\&WS) results in more subtle performance fluctuations and less frequent drift detector triggers. Other strategies mainly trigger the drift detector in the beginning and after abrupt drift. 

To evaluate the effect of the drift detector, we replaced it with five pre-set drift points: one after the abrupt drift and four after the mid-gradual drifts. Fig. \ref{fig:drifts}-b compares the accuracy scores for H2O with the drift detector (full line) and the pre-sets (dashed). Although there are minor differences, e.g. faster recovery with D\&RS, overall the relative performance of adaptation strategies does not change.

\begin{figure}[!t]
    \centering
    \includegraphics[trim=150 20 120 0,width=2.5in]{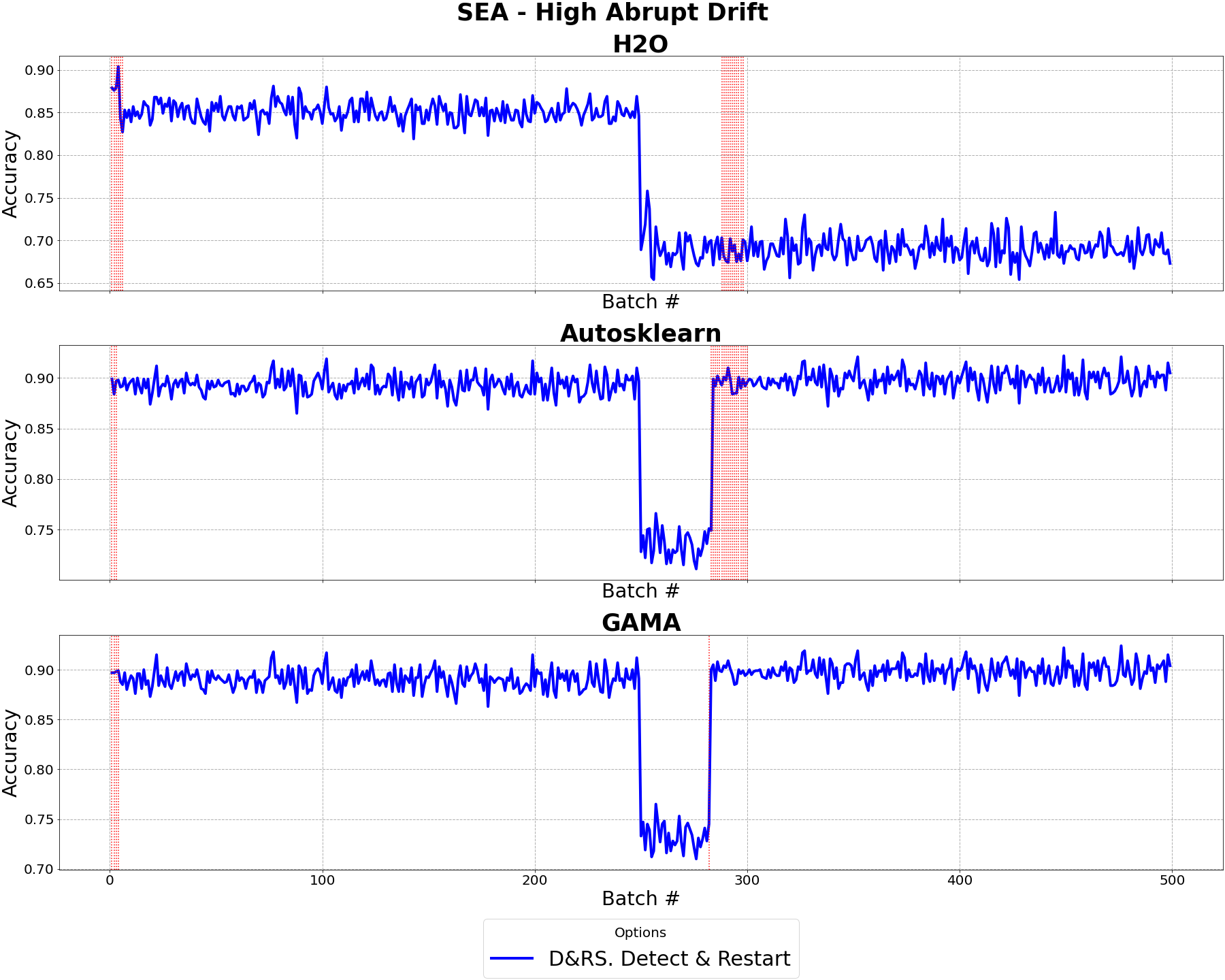}
    \caption{Accuracy across batches with drift detection points for SEA - High abrupt data stream on D\&RS strategy.}
    \label{fig:automl}
\end{figure}

%NEW 
Finally, Fig. \ref{fig:automl} marks the detected drift points (in red) for each library for the high abrupt drift data stream with D\&RS. Drift is always detected after the sudden drift point at batch 250. Autosklearn and GAMA recover after the drift is detected and pipelines are re-optimized, but not H2O. Hence, a lack of drift detection is not the reason for H2O's performance after drift. Similar results were found for the abrupt and mixed drift data streams. This suggests that the linear model (GLM) used in H2O's stacking process is not adapting well to the changes in data distribution. Replacing this with a more adaptive stacker (e.g. GBM) may resolve this. H2O is also still a competitive option for gradual drift data streams with D\&I chosen as the adaptation strategy.
%except for D\&W for Autosklearn which may be due to the long term memory that is a result of the modification choices. 

\begin{figure}[!t]
\centering
\includegraphics[trim=150 20 120 0,width=2.5in]{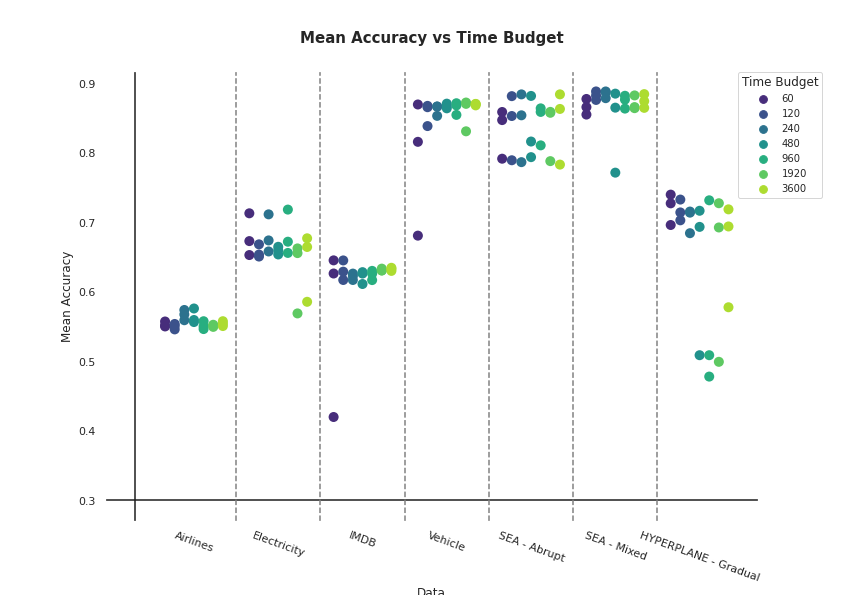}
 \caption{Mean accuracy for all data streams and libraries with D\&RS on time budgets between \SI{60} and \SI{3600} seconds.}
 \label{fig:time}
\end{figure}

\subsection{Effect of the time budget}
When we choose to rerun an AutoML technique (e.g. in D\&RS and PRS), we can do so in parallel and keep using the previous pipelines until the new pipelines are optimized to new data. Still, we may want to limit the time budget given to the AutoML system so that our method reacts more quickly to concept drift.
Fig. \ref{fig:time} shows the accuracy of all libraries with the D\&RS strategy on each data stream, for time budgets varying from \SI{60}{sec} to \SI{3600}{sec}. Smaller budgets do affect performance, but overall most methods work well under stricter constraints. For data streams with a larger feature space (i.e. IMDB, Vehicle) the time budget has a more visible effect compared to artificial and lower-dimensional data streams. Blast, which trains multiple models in an ensemble, has similar accuracy and time performance. An advantage of the AutoML methods is that the time budget can be restricted without much loss in accuracy, which allows various online learning settings. An interesting research question would be to examine the relationship between the optimal time budget and the characteristics of the data stream.

\section{Conclusions}\label{sec:conclusions}
The main goal of this study was to gain a deeper understanding on how current AutoML methods are affected by different types of concept drift, and how they could be adapted to become more robust. To this aim, we proposed six adaptation strategies that can be generally combined with AutoML techniques, integrated them with some of the most well-known AutoML approaches, and evaluated them on both artificial and real-world datasets with different types of concept drift.

We found that these strategies effectively allow AutoML techniques to recover from concept drift, and rival or outperform popular or online learning techniques such as Oza Bagging and BLAST. The comparisons between different AutoML techniques show that both Bayesian optimization and evolutionary approaches can be adapted to handle concept drift well, given an appropriate adaptation strategy and forgetting mechanism. Similar to previous studies, we find that different drift characteristics affect learning algorithms in different ways, and that different adaptation strategies may be needed to optimally deal with them.

On large datasets, re-optimizing pipelines after drift is detected works generally well. Evolutionary AutoML methods can so do even faster by evolving the previous best pipelines. Simply retraining pipelines on the most recent data after drift is detected, without re-optimizing the pipelines themselves, also works well if the concept drift is not too large and the pipelines are retrained frequently enough. Depending on the application, the additional computational time can be mitigated by decreasing the batch size, the optimization time budget, and warm-starting.

In all, this study contributes a set of promising adaptation strategies as well as an extensive empirical evaluation of these strategies, so that informed design choices can be made on how to adapt AutoML techniques to settings with evolving data. It also shows that there is ample room to improve existing AutoML systems, and even to design entirely new AutoML systems that naturally adapt to concept drift. We hope that this study will instigate further research into AutoML methods that adapt effortlessly to evolving data.

%For instance, AutoML techniques based on Bayesian Optimization could benefit from novel warm-starting approaches, and AutoML techniques that stack multiple pipelines require more adaptive stacking techniques. 

%Future: statistical significance - reruns?
% There is ample room for improvemenrt, e.g. H2O stacking or AutoSKlearn warm-starting

\section*{Acknowledgments}
We would like to thank Erin Ledell, Matthias Feurer and Pieter Gijsbers for their advice on adapting their AutoML systems. This work is supported by the Dutch Science Foundation (NWO) grant DACCOMPLI (nr. 628.011.022) and partially supported by TAILOR, a project funded by EU Horizon 2020 research and innovation programme (under GA No 952215).

\bibliographystyle{IEEEtranS}
\bibliography{References.bib}

\vspace{-1.3cm}
\begin{IEEEbiography}[{\includegraphics[width=1in,height=1.20in,clip,keepaspectratio]{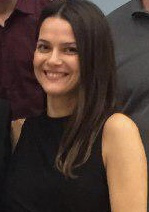}}]{Bilge Celik}
is a PhD researcher in the  Department of Mathematics and Computer Science at Eindhoven University of Technology. She received her M.S. and B.A. degrees from Middle East Technical University. Her research interests include automated machine learning, data stream challenges and automation in adaptive learning.
\end{IEEEbiography}
\vspace{-1.5cm}
\begin{IEEEbiography}[{\includegraphics[width=0.9in,height=1.25in,clip,keepaspectratio]{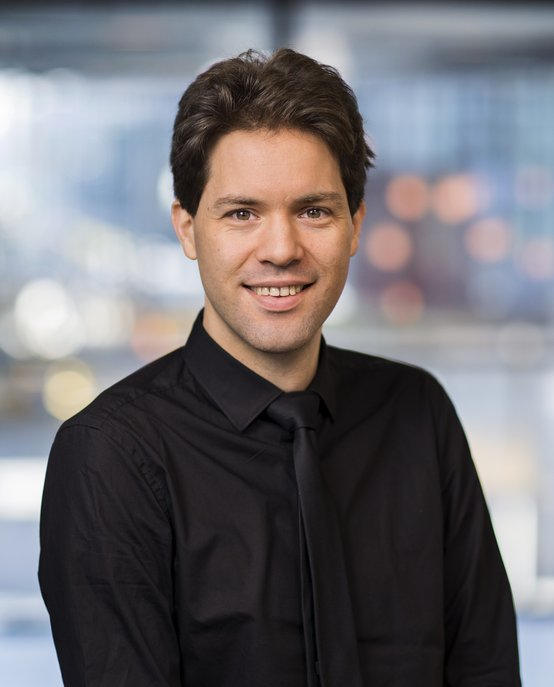}}]{Joaquin Vanschoren}
focuses his research on automated machine learning (AutoML) and meta-learning (learning to learn). He founded the OpenML project (openml.org) and co-organizes the AutoML and meta-learning workshops at ICML and NeurIPS. He co-presented tutorials at NeurIPS and AAAI, and co-authored the book 'Automated Machine Learning' (Springer, 2019). He’s a founding member of ELLIS and CLAIRE, and action editor at JMLR.
\end{IEEEbiography}

%\newpage

%vfill


\end{document}